\documentclass[sigconf]{aamas}

\usepackage{balance}
\usepackage{subcaption}
\usepackage{enumitem}
\usepackage{tabularray}
\usepackage{siunitx}

\usepackage{amsmath} 
\usepackage[linesnumbered,ruled,vlined,noend]{algorithm2e}
\usepackage{graphicx} 

\usepackage{float} 

\usepackage{algpseudocode} 
\usepackage{xcolor} 
\usepackage{boxedminipage}
\usepackage{multirow}
\usepackage{lipsum}
\usepackage{url}
\usepackage{xcolor}

\setcopyright{none}
\renewcommand\footnotetextcopyrightpermission[1]{} 
\title[AAMAS-2026 Formatting Instructions]{MeCo: Enhancing LLM-Empowered Multi-Robot Collaboration via Similar Task Memoization}

\author{Baiqing Wang}
\affiliation{
  \institution{Northwestern Polytechnical University}
  \city{Xi'an}
  \country{China}}
\email{wbq@mail.nwpu.edu.cn}

\author{Helei Cui}
\authornote{Corresponding author.}
\affiliation{
  \institution{Northwestern Polytechnical University}
  \city{Xi'an}
  \country{China}}
\email{chl@nwpu.edu.cn}

\author{Bo Zhang}
\affiliation{
  \institution{Northwestern Polytechnical University}
  \city{Xi'an}
  \country{China}}
\email{bo.zhang@mail.nwpu.edu.cn}

\author{Xiaolong Zheng}
\affiliation{
  \institution{Beijing University of Posts and Telecommunications}
  \city{Beijing}
  \country{China}}
\email{zhengxiaolong@bupt.edu.cn}

\author{Bin Guo}
\affiliation{
  \institution{Northwestern Polytechnical University}
  \city{Xi'an}
  \country{China}}
\email{guob@nwpu.edu.cn}

\author{Zhiwen Yu}
\affiliation{
  \institution{Harbin Engineering University \&\\ \mbox{Northwestern Polytechnical University}}
  \city{Harbin}
  \country{China}}
\email{zhiwenyu@nwpu.edu.cn}

\begin{abstract}

Multi-robot systems have been widely deployed in real-world applications, providing significant improvements in efficiency and reductions in labor costs. However, most existing multi-robot collaboration methods rely on extensive task-specific training, which limits their adaptability to new or diverse scenarios. Recent research leverages the language understanding and reasoning capabilities of large language models (LLMs) to enable more flexible collaboration without specialized training. Yet, current LLM-empowered approaches remain inefficient: when confronted with identical or similar tasks, they must replan from scratch because they omit task-level similarities. To address this limitation, we propose MeCo, a similarity-aware multi-robot collaboration framework that applies the principle of ``cache and reuse'' (a.k.a., memoization) to reduce redundant computation. Unlike simple task repetition, identifying and reusing solutions for similar but not identical tasks is far more challenging, particularly in multi-robot settings. To this end, MeCo introduces a new similarity testing method that retrieves previously solved tasks with high relevance, enabling effective plan reuse without re-invoking LLMs. Furthermore, we present MeCoBench, the first benchmark designed to evaluate performance on similar-task collaboration scenarios. Experimental results show that MeCo substantially reduces planning costs and improves success rates compared with state-of-the-art approaches.
\end{abstract}


\newcommand{\BibTeX}{\rm B\kern-.05em{\sc i\kern-.025em b}\kern-.08em\TeX}

\begin{document}

\pagestyle{fancy}
\fancyhead{}

\maketitle

\begingroup
\renewcommand\thefootnote{}
\footnotetext{
  \vspace{-0.4\baselineskip}
  \noindent\rule{\linewidth}{0.4pt}\par
  \vspace{0.6\baselineskip}
  \noindent Preprint Version. To be published in \emph{Proc.\ of the 25th International Conference on Autonomous Agents and Multiagent Systems (AAMAS~2026).}
}
\addtocounter{footnote}{-1}
\endgroup

\begin{figure*}[t]
    \centering
    \includegraphics[width=0.96\linewidth]{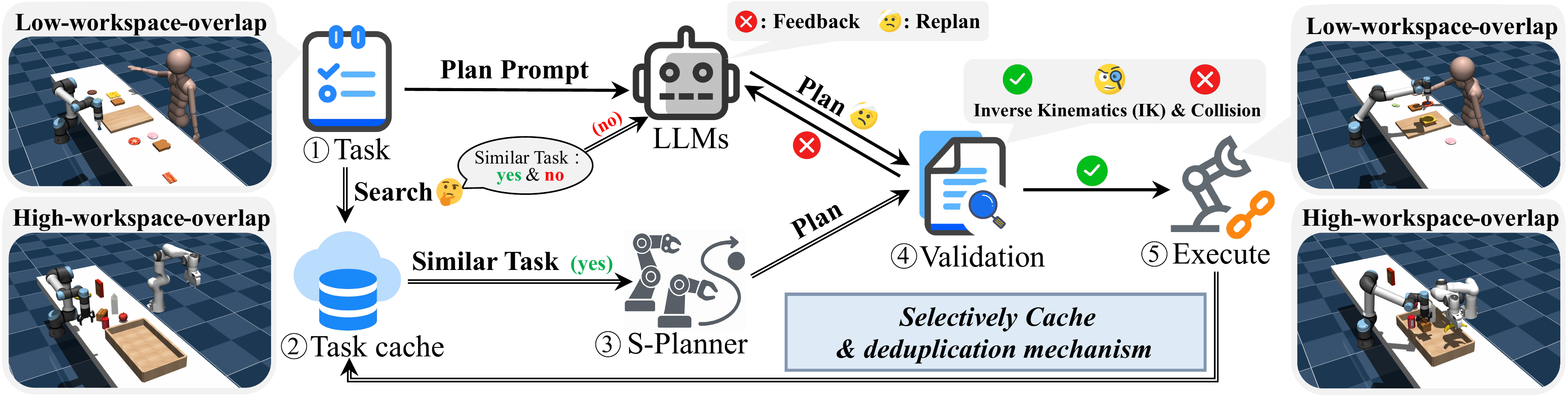}
    \caption{Workflow of MeCo. The arrow $\rightarrow$ shows the LLM-empowered planning; the arrow $\Rightarrow$ shows the MeCo extensions.}
    \label{MeCo}
\end{figure*}

\section{Introduction}

Multi-robot systems have appeared in many real-world scenarios, like automatic manufacturing~\cite{lai2025roboballet,goldberg2019robots,wete2024multi}, smart housekeeping~\cite{mao2024multimodal,mahler2019learning,leidner2016robotic}, and intelligent healthcare~\cite{lee2024levels,schmidgall2024general,agrawal2023shelfhelp}. 
They significantly improve production efficiency and reduce labor costs. 
Current methods, including teach programming~\cite{mehta2016teach}, reinforcement learning (RL)~\cite{zhang2021multi,ju2022transferring}, and imitation learning (IL)~\cite{ho2016generative}, rely on fixed programming or prolonged training to generate collision-free trajectories for specific tasks, limiting their adaptability across diverse tasks. 
By leveraging the superior text comprehension, reasoning, and generation capabilities of large language models (LLMs)~\cite{radford2019language, brown2020language, raffel2020exploring,chen2025future}, multi-robot systems achieve flexible handling of diverse coordination scenarios without any task-specific training.

In this direction, a number of LLM-empowered multi-robot collaboration frameworks are proposed, e.g., RoCo~\cite{mandi2024roco}, EMOS~\cite{chen2024emos}, and ReCA~\cite{wan2025reca}. 
They assign robots to LLM agents, which plan tasks either in a centralized manner or via dialogue-based coordination.
With the ability to capture common-sense knowledge~\cite{zhao2023large}, LLMs empower agents to provide robots with intelligent task-decomposition and flexible decision-making.
Despite promising, the above-mentioned studies all treat tasks as isolated ones and overlook the inter-task relationships.

Similar tasks often appear in real practice.
Examples include parcel sorting~\cite{amazonUniversalModel} and product categorization~\cite{amazonPinchgraspingRobot} in assembly lines, as well as floor sweeping, sandwich making, and grocery packing in home automation~\cite{mao2024multimodal}. 
While task-specific elements (e.g., target object types and positions) may vary to some extent, the core objectives (e.g., sorting parcels into containers or sweeping items into baskets) and inherent constraints (e.g., robot quantity and fixed resources like containers) remain unchanged. 
However, existing LLM-empowered methods still re-invoke LLM agents when encountering similar tasks, leading to unnecessary costs, i.e., \mbox{token} consumption and planning time. 
\mbox{Although} some approaches consider reducing tokens, such as HMAS-2~\cite{chen2024scalable}, S-ATLAS~\cite{wang2024probabilistically}, and ReAd~\cite{zhang2024towards}, they can hardly cut the planning costs caused by redundant replanning for similar tasks. 

In light of these, we propose MeCo, a similarity-aware multi-robot collaboration framework that leverages task-level similarity for efficient planning.
However, compared with dealing with exactly the same tasks, finding and leveraging the similarity between two tasks is more challenging.
In response, a new similarity testing method is designed to identify previously solved tasks with high relevance, enabling the reuse of existing plans without \mbox{re-invoking} the LLMs.
As shown in Figure~\ref{MeCo}, upon receiving a task, MeCo first searches the task cache for a similar task. 
If no similar task is found, it invokes LLMs for planning, just like the previous work. 
If an identical or similar task is matched, we further involve a similar motion planner (\mbox{S-Planner}), which generates a plan for the current task by referencing the similar task without relying on LLMs. 
If S-Planner fails, the failure reason is fed back to the LLMs, which continue planning from the failed step instead of restarting the entire process. 
Once planning is successful, MeCo selectively stores this task in the task cache. 
To facilitate evaluation, we further introduce a variant of RoCoBench~\cite{mandi2024roco} for similar tasks, called MeCoBench. 
Experiments on MeCoBench show that our proposed design achieves an overall improvement of around 30\% in success rate, about 55\% saving in planning time, and up to 70\% reduction in token consumption, compared with the state-of-the-art design.

\textbf{\textit{Contribution.}} To the best of our knowledge, we are the first to explore the problem of similar tasks in multi-robot collaboration. The primary contributions of our work are outlined as follows:

\begin{itemize}[left=3pt, itemsep=0pt, topsep=3pt, parsep=0pt]
\item We propose MeCo, a similarity-aware multi-robot collaboration framework that leverages task-level similarity to enhance task planning powered by LLMs.

\item We present a similarity testing method that employs multiple strategies to accurately evaluate task similarity, based on the degree of workspace overlap among robots.

\item We design S-Planner, a similar motion planner that references previously executed similar tasks to plan new ones, thereby avoiding repeated LLM invocations.

\setcounter{footnote}{0}
\item We introduce MeCoBench, an extended benchmark designed to evaluate performance on similar tasks. Experimental results demonstrate that MeCo significantly reduces planning costs while improving task success rates. We open-source our code.\footnote{\url{https://github.com/TomWang-NPU/MeCo}}

\end{itemize}

\section{Related work} \label{sec:1}

\textbf{LLMs for Robotics.} Early prior work uses LLMs to select skill primitives and complete robotic tasks, such as SayCan~\cite{ahn2022can} and Inner Monologue~\cite{huang2023inner}. Later work utilizes LLMs to generate code-formatted robotic policies, such as ProgGPT~\cite{singh2023progprompt}, Code-As-Policy~\cite{liang2023code}, and ELLMER~\cite{mon2025embodied}. 
These works primarily focus on single-robot setups. Recent advancements extend LLM applications to multi-robot scenarios. For instance, Zhang \textit{et al.}~\cite{zhang2023building} utilize LLMs as planners to facilitate agent collaboration in dialogue-driven tasks. Zhao~\textit{et~al.}~\cite{mandi2024roco} propose an LLM-based multi-robot framework (RoCo), assigning each robot to an LLM agent for task collaboration in a dialogue style. 
Wan \textit{et al.}~\cite{wan2025reca} introduce a characterization and co-design framework (ReCA), which improves task efficiency and system scalability in LLM-based multi-robot collaboration, while also reducing token usage. 
Wang \textit{et al.}~\cite{wang2024probabilistically} develop an LLM-based distributed multi-robot planner using conformal prediction, which achieves the user-specified success rate while minimizing token usage. 
Chen \textit{et al.}~\cite{chen2024scalable} compare four LLM planning frameworks in multi-robot coordination to achieve higher token efficiency. 
While prior work achieves effective multi-robot coordination via LLMs and proposes token cost and planning time reduction methods, scenarios containing similar tasks still require repeated LLMs planning, resulting in unnecessary token costs and planning time. 

\textbf{Learning-based methods for Robotics.} Learning-based methods, such as reinforcement learning (RL)~\cite{wu2023learning,pal2025together,yu2025adaptive} and imitation learning (IL)~\cite{ho2016generative,seo2025hierarchical,xu2019acousticid}, have been applied to multi-robot collaboration. These methods typically require task-specific training to allocate actions and plan paths. For example, Xie \textit{et al.}~\cite{xie2020deep} propose a deep imitation learning framework for robotic bimanual manipulation in a continuous state-action space. Cui \textit{et al.}~\cite{cui2024task} introduce an actor-critic-based deep reinforcement learning framework for closed-chain manipulation in dual-arm robotic systems. Although learning-based methods have achieved success, in multi-robot collaboration, they require modifications or retraining when tasks change, making them unsuitable for scenarios with similar tasks.

\textbf{Programming methods for Robotics.} Programming methods for robotics are widely used in intelligent manufacturing, such as part assembly on industrial production lines and fixed-item sorting. These methods typically adopt two paradigms: offline programming~\cite{visualcomponentsRobotOffline} for predefined trajectories and teach pendant programming~\cite{mehta2016teach} for manual path recording. However, their reliance on static task assumptions limits applicability in environments with evolving operational demands.

\begin{figure*}[t]
    \centering
    \begin{subfigure}[b]{0.46\textwidth}
        \centering
        \includegraphics[width=\linewidth]{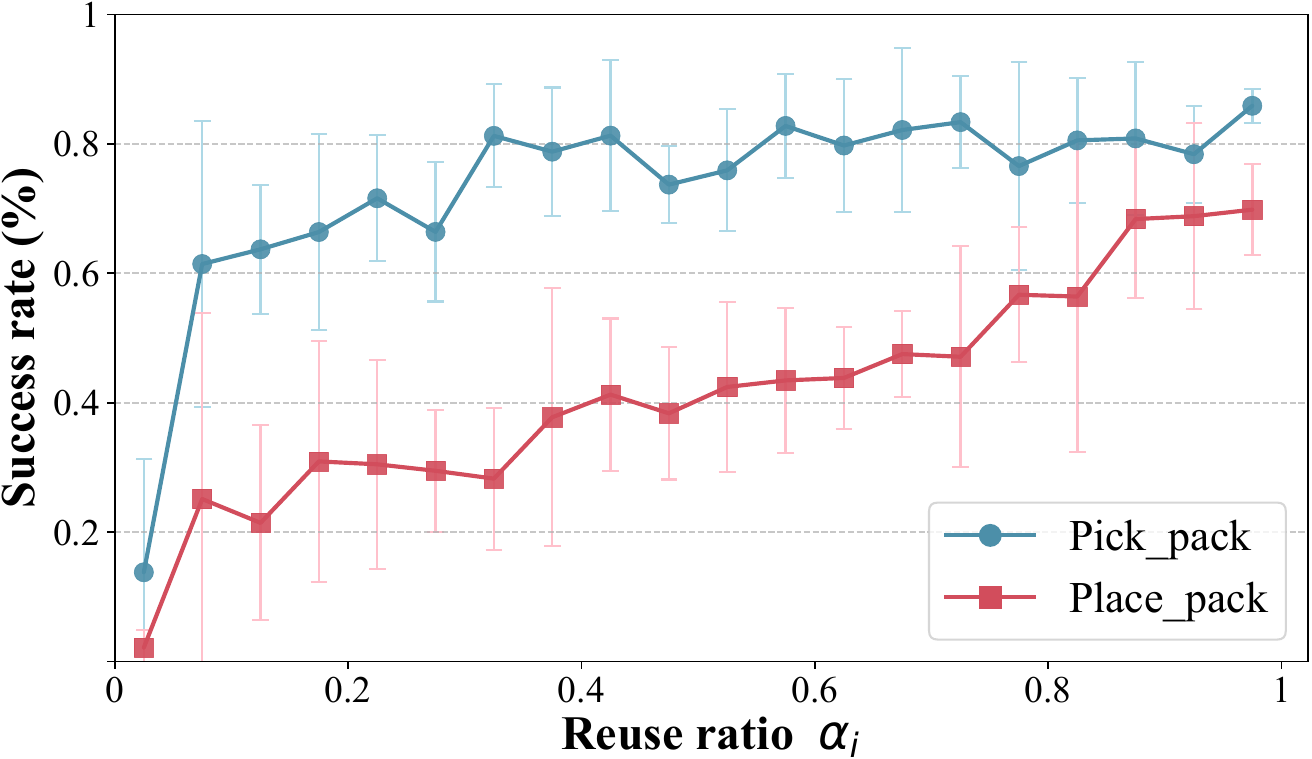}
        \caption{Pack Grocery}
        \label{alpha_1}
    \end{subfigure}
    \hfill
    \begin{subfigure}[b]{0.46\textwidth}
        \centering
        \includegraphics[width=\linewidth]{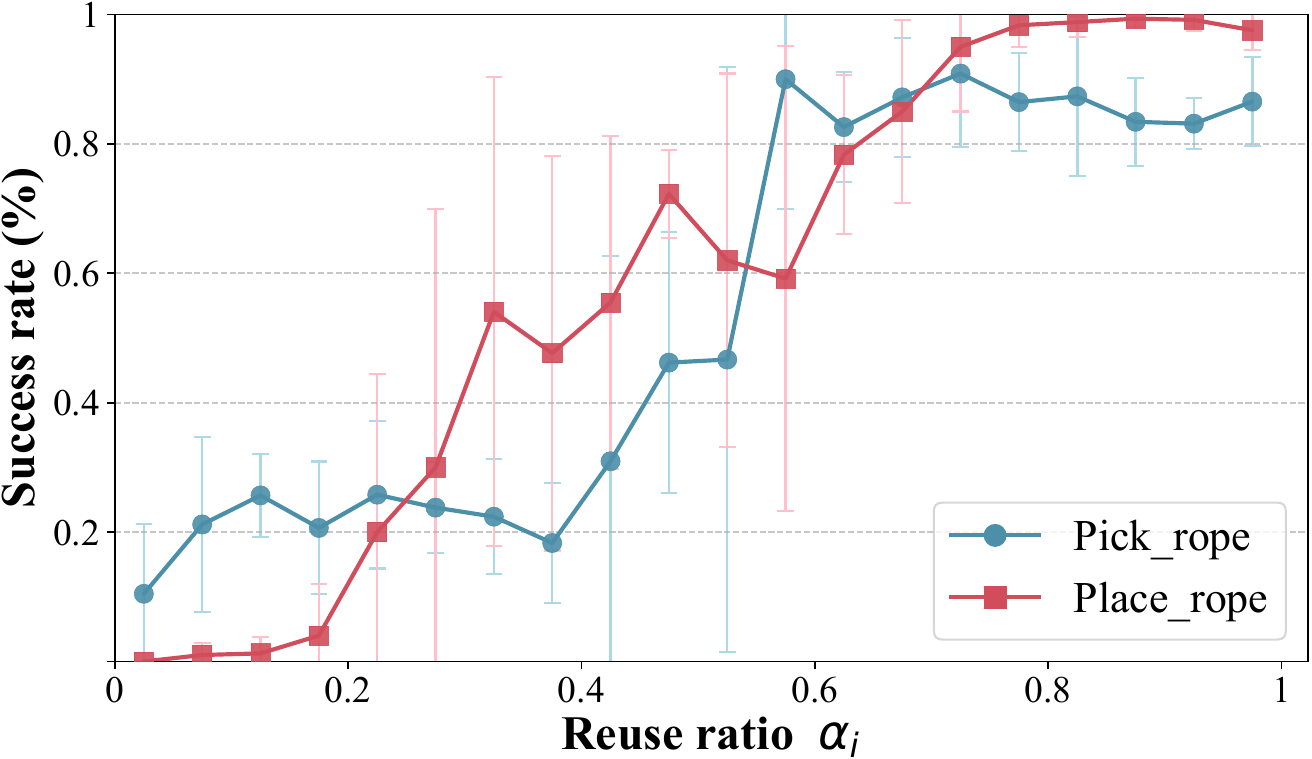}
        \caption{Move Rope}
        \label{alpha_2}
    \end{subfigure}
    \caption{The value of $\alpha_i$ affects the success rate of planning the current task based on similar tasks. We test the ``Pick'' and ``Place'' subtasks of the \textit{Pack Grocery} and \textit{Move Rope} tasks under different intervals, with each setting tested 100 times on average.}
    \label{alpha}
\end{figure*}

\section{Problem statement} \label{sec:2}

\subsection{LLM-empowered planning}

We use LLMs to generate 3D waypoint paths that guide low-level motion planning. This practice is common, such as in RoCo~\cite{mandi2024roco}, ReAd~\cite{zhang2024towards}, AO-Planner~\cite{chen2025affordances}, and \cite{lin2025think,ji2025collision}.
As shown in the single-line section of Figure~\ref{MeCo}, the task description is provided to the LLMs in text form. LLMs decompose the task into multiple subtasks. For each subtask, LLMs output 3D waypoint paths in the robots' workspace to guide the low-level motion planner. The generated 3D waypoints undergo inverse kinematics (IK) and collision checking to ensure their validity. If all waypoints pass validation, the low-level motion planner is invoked to plan and execute the trajectory. Otherwise, the failure reason is fed back to the LLMs for replanning.

\subsection{Definition of tasks}
A task is defined as a tuple \( T = (C, E) \). The task content \( C \) is a text-based specification of execution requirements. The environmental elements \( E \) define the task space \( \Omega \) through \( E = (R, O, D, K) \), where \( R = \{\{ r_i = (\tau_i, n_i, W_i) \}
\}_{i=1}^{N} \) represents robots, with each \( r_i \) comprising type \( \tau_i \), quantity \( n_i \), and workspace \( W_i \) (\( W_i \in \Omega \)); \( O = \{o_j= (p_j, \gamma_j\}_{j=1}^{M} \) denotes target objects, where each \( o_j \) includes position \( p_j \) and category \( \gamma_j \); \( D \) captures obstacle distributions; and \( S \) encodes task-specific constraints, such as recipes or ordering~constraints. 

For demonstration purposes, we focus on the six tasks in RoCoBench~\cite{mandi2024roco}, a state-of-the-art benchmark of multi-robot collaboration scenarios, which is widely used in previous studies~\cite{chen2024internet,zhang2024towards,lin2025think}.
RoCoBench consists of six multi-robot collaboration tasks in a tabletop manipulation environment, including \textit{Move Rope}, \textit{Pack Grocery}, \textit{Make Sandwich}, \textit{Sort Cubes}, \textit{Arrange Cabinet}, and \textit{Sweep Floor}. Each task imposes specific workspace constraints on the robots involved. We categorize tasks into two groups based on robots' workspace overlap levels: \textbf{low-workspace-overlap tasks} and \textbf{high-workspace-overlap tasks}. Tasks such as \textit{Sweep Floor}, \textit{Make Sandwich}, \textit{Sort Cubes}, and \textit{Arrange Cabinet} fall under low-workspace-overlap tasks, while \textit{Pack Grocery} and \textit{Move Rope} belong to high-workspace-overlap tasks.

\section{Our proposed design} \label{sec:3}

\subsection{Overview}
As shown in Figure~\ref{MeCo}, after each successful planning, MeCo selectively stores the task plan in the task cache for future reference. When planning a new task, it first searches the task cache for a similar task that meets the current requirements. If no similar task is found, MeCo calls LLMs and uses the task description as the prompt for planning. 
If a similar task is matched, it then invokes the similar motion planner, namely S-Planner, which generates a plan for the current task efficiently by referencing the similar task without relying on LLMs. 
S-Planner or LLMs decompose the task into multiple subtasks. For each subtask, they output 3D waypoint paths to guide the low-level motion planner. 
The generated plan undergoes a set of validations, such as inverse kinematics (IK) and collision checking. If validation fails, the failure reason is fed back to LLMs for replanning. Specifically, if S-Planner fails to generate a valid plan, LLMs continue planning from the failed step instead of restarting from scratch. 
Once all validations are passed, the final motion trajectories for all robots are planned by the low-level motion planner and executed in the environment.
The detailed workflow is provided in the appendix.

\begin{figure*}[t] 
    \centering
    \begin{subfigure}{0.98\textwidth}
        \centering
        \includegraphics[width=\linewidth]{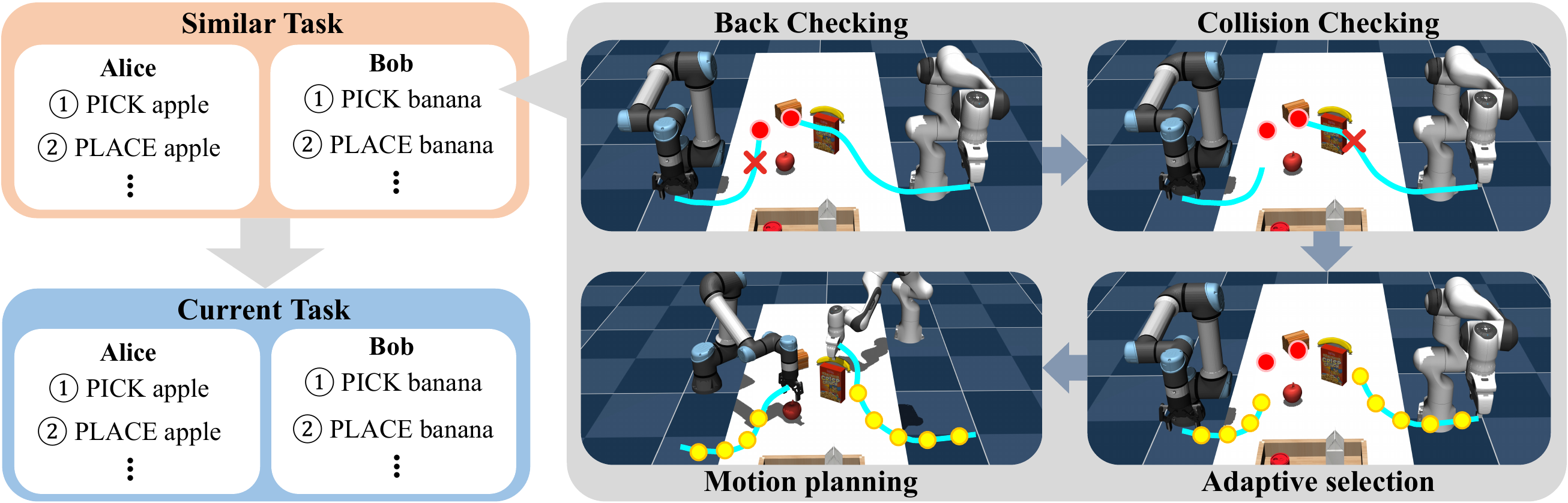}
    \end{subfigure}
    \caption{Workflow of S-Planner in the high-workspace-overlap mode. 
    The left side illustrates how tasks are decomposed and assigned at the high level, 
    while the right side shows how motion planning is conducted at the low level. 
    The blue curves represent the reference trajectory, 
    the red points mark the target positions in the similar task;
    `Alice' and `Bob' refer to robots.}
    \label{S-Planner}
\end{figure*}

\subsection{Definition of similar tasks}\label{sec:A}
In the search phase of MeCo, it needs to identify whether there are any tasks in the task cache that are similar to the current task and select the most similar task for use in subsequent stages. We have defined task similarity criteria separately for high-workspace-overlap tasks and low-workspace-overlap tasks.

\textbf{For low-workspace-overlap tasks,} the task space $\Omega$ is partitioned into a finite set of disjoint work areas
$\mathcal{A} = \{A_1, \ldots, A_M\}$,such as the left and right regions of cutting boards in \textit{Make Sandwich} or the regions between panels in \textit{Sort Cubes}. 
The historical task is denoted as $T_{\mathrm{ref}} = (C, E_{\mathrm{ref}})$, and the current task as $T_{\mathrm{cur}} = (C, E_{\mathrm{cur}})$. Both tasks share the same $C, R, D,$ and $S$.
$T_{\mathrm{cur}}$ and $T_{\mathrm{ref}}$ are considered similar if there exists a point mapping, as shown in Eq.~\eqref{eq1}, that ensures each target object undergoes only restricted geometric transformations within the same region, without altering the cross-region interaction structure.
\begin{equation}
\Pi : \Omega \to \Omega, \; \text{s.t. } \forall j,\; p^{\mathrm{cur}}_j = \Pi(p^{\mathrm{ref}}_j),\;
p^{\mathrm{ref}}_j \in A_m \Rightarrow p^{\mathrm{cur}}_j \in A_m,
\label{eq1}
\end{equation}
where $p^{\mathrm{ref}}_j$ and $p^{\mathrm{cur}}_j$ denote the position of the $j$-th target object in the historical and current tasks, respectively.

\textbf{For high-workspace-overlap tasks,} there is a higher risk of collisions. Therefore, it is necessary to further consider the reusability of historical trajectories in the current task environment. The historical task $T_{\mathrm{ref}} = (C, E_{\mathrm{ref}})$ and the current task $T_{\mathrm{cur}} = (C, E_{\mathrm{cur}})$ share the same $C, R, D,$ and $S$. A task is decomposed into several subtasks, each associated with a reference trajectory of the robot end-effector on the horizontal plane. 
For the $i$-th subtask, we define:
\begin{itemize}[leftmargin=*]
    \item $\Delta x_i^{\text{goal}}$: the horizontal distance from the target object to its goal position in the current task;
    \item $\Delta x_i^{\text{ref}}$: the horizontal length of the reference trajectory from the historical task that is reusable in the current environment.
\end{itemize}
Accordingly, the reuse ratio $\alpha_i$ is defined as:
\begin{equation}
\alpha_i = \frac{\Delta x_i^{\text{ref}}}{\Delta x_i^{\text{goal}}},
\end{equation}
Furthermore, the overall task similarity metric is defined as the average reuse ratio across all subtasks:
\begin{equation}
\alpha = \frac{1}{n} \sum_{i=1}^{n} \alpha_i,
\end{equation}
where $n$ denotes the number of subtasks in the historical task plan.

Let $p$ denote the unit failure rate of reusing historical trajectories in the current environment. 
If the reusable trajectory length is $\Delta x_i^{\text{ref}}$, its success rate is $(1-p)^{\Delta x_i^{\text{ref}}}$. 
For the remaining trajectory length $\Delta x_i^{\text{goal}} - \Delta x_i^{\text{ref}}$, a planner generates new trajectories. 
Let $\tilde{p}$ denote the unit failure rate of replanning. Since replanning is more difficult, $\tilde{p} > p$. 
The success rate of the replanned segment is therefore $(1-\tilde{p})^{\Delta x_i^{\text{goal}} - \Delta x_i^{\text{ref}}}$. 
By multiplying the two terms and substituting $\Delta x_i^{\text{final}} = \alpha_i \Delta x_i^{\text{goal}}$, we obtain the lower bound of the subtask success probability:
\begin{equation}
P_{\text{succ},i} \geq (1-\tilde{p})^{\Delta x_i^{\text{goal}}} \cdot \exp\!\left( \kappa \cdot \alpha_i \cdot \Delta x_i^{\text{goal}} \right),
\label{eq:succ-prob}
\end{equation}
where $\kappa = \ln \frac{1-p}{1-\tilde{p}} > 0$. 
This inequality shows that $P_{\text{succ},i}$ increases monotonically with $\alpha_i$.
Furthermore, we conducted experiments on two sub-tasks, "Pick" and "Place", within the high-workspace-overlap tasks of \textit{Pack Grocery} and \textit{Move Rope}, as shown in Figure~\ref{alpha}. The results indicate that as $\alpha_i$ increases, the success rate of sub-tasks improves. Therefore, we use $\alpha$ as the indicator for determining the similarity of high-workspace-overlap tasks.

MeCo can select an appropriate threshold $\tau$ based on its requirements. If $\alpha < \tau$, the tasks are considered dissimilar; otherwise, they are considered similar. For lower values of $\tau$, MeCo relaxes the similarity criteria for task selection, resulting in a lower success rate. In contrast, a higher $\tau$ increases the stringency of task selection, leading to a higher success rate. We found that for the \textit{Pack Grocery}, $\alpha$ = 0.85, and for the \textit{Move Rope}, $\alpha$ = 0.7, the success rate shows little fluctuation while maintaining a high level of performance. Therefore, we choose 0.85 and 0.7 as the thresholds for these two tasks in our subsequent experiments.

\subsection{Similar motion planning}\label{sec:B}

We now introduce the S-Planner for similar motion planning. Briefly, S-Planner plans the current task by referring to the plans of similar tasks. It operates in two modes: one for tasks with high workspace overlap and the other for tasks with low workspace~overlap. 

\textbf{Mode for low-workspace-overlap tasks.} By referring to the similar task $T_{\text{ref}}$, S-Planner decomposes the current task $T_{\text{cur}}$ into multiple subtasks. 
Each subtask corresponds to an action sequence $\{a_1, a_2, \ldots, a_n\}$, where $n$ denotes the number of robots. 
At the high level, S-Planner inherits the action sequence from the similar task and applies it to the current task:
$a_i^{\text{cur}} = a_i^{\text{ref}}, \quad \forall i \in \{1, \ldots, n\}.$
At the low level of motion planning, we define the target pose difference based on the target pose $g_s$:
\begin{equation}
\Delta(g_s^{\text{ref}}, g_s^{\text{cur}}) = \| g_s^{\text{ref}} - g_s^{\text{cur}} \|,
\end{equation}
where $\|\cdot\|$ is a pose difference metric that combines translational and rotational distances.
We then define the decision function as
\begin{equation}
\delta_s = \mathbf{1}\!\left[ \Delta(g_s^{\text{ref}}, g_s^{\text{cur}}) \leq \varepsilon \right],
\end{equation}
where $\varepsilon > 0$ is the tolerance threshold.

\begin{itemize}[leftmargin=*]
    \item If $\delta_s = 1$: the target pose remains unchanged, and the trajectory is reused directly.
    \item If $\delta_s = 0$: the target pose changes, and the RRT-based motion planner is invoked to generate a new trajectory.
\end{itemize}

\textbf{Mode for high-workspace-overlap tasks.} At a high level, S\mbox{-}Planner follows the same approach as the mode for low-workspace-overlap tasks, inheriting the action sequence from the similar task and applies it to the current task:
$a_i^{\text{cur}} = a_i^{\text{ref}}, \quad \forall i \in \{1, \ldots, n\}.$ However, since high-workspace-overlap tasks have a higher risk of collision, the low-level planning method used for low-workspace-overlap tasks is not applicable.
In particular, the motion trajectory in the similar task plan may not always be applicable to the current task. For instance, the positions of non-target objects in the current task may change and overlap with the reference trajectory. Therefore, it is essential to identify the valid reference segments of the similar task trajectory, which involves two steps: back checking and collision checking, as shown in Figure~\ref{S-Planner}.

Let the joint-space sequence generated by the historical plan be $Q = \{ q_s \}_{s=0}^N,$ where each $q_s$ is a set of joint angles. Using forward kinematics, $\text{FK}: \mathbb{R}^{d_q} \to \mathbb{R}^3,$ the corresponding end-effector trajectory is
$R = \{ P_s = (x_s, y_s, z_s) \}_{s=0}^N.$ The first step, back checking, prevents unnecessary backtracking during task execution.
Specifically, if the target position \( P_c \) of the current task is located in the middle of the reference path $R$, the path is truncated to obtain $R$. This ensures that the trajectory does not return to previously visited regions. Accordingly, the truncated path is defined as follows:  
\begin{equation}
R_b = \left( P'_s = (x_s, y_s, z_s)\right)_{s:\; d(x_s - x_c) \geq 0} ,
\end{equation}
where \( \text{R}_b \subseteq \text{R} \), and $d := \text{sign}(x_N - x_0) \in \{+1, -1\}$ denoting the monotonic direction indicator. Next, collision checking is performed on \( \text{R}_b \) to verify whether the reference motion trajectory is collision-free in the current task environment. If collisions are detected, \( \text{R}_b \) is truncated before the collision point to obtain \( \text{R}_\text{final} \), where \( \text{R}_\text{final} \subseteq \text{R}_b \), ensuring a feasible reference trajectory.  

During trajectory planning, S-Planner adaptively selects key sampling points from \( \text{R}_\text{final} \) to reduce redundancy while preserving critical information. The highest point along the z-axis, \( P_\text{peak} \), is identified as the central key point. Using an expansion coefficient \( \eta\), the expansion step size is computed as \( \Delta = \lfloor \eta \cdot N \rfloor \), where \( N \) is the total number of points in the path. Based on \( \Delta \), expansion continues in both directions within \( \text{R}_\text{final} \) until reaching the endpoints on both sides, forming the point sets \( K_\text{left} \) and \( K_\text{right} \). Thus, the final set of key sampling points is defined as  \(K = \{ P_{\text{peak}} \} \cup K_{\text{left}} \cup K_{\text{right}} .\) Finally, \( K \) is provided as an important reference to the RRT-based motion planner for generating motion trajectories for all robots.  

\subsection{Continuous planning}
Although S-Planner can plan by referencing previous task plans without calling LLMs, its effectiveness is influenced by the similarity between the previous and current tasks. If S-Planner fails to generate a valid plan, MeCo reassigns the task to LLMs for replanning. Since S-Planner failures typically occur at intermediate steps, having LLMs replan from scratch would render the previous S-Planner planning efforts ineffective. To address it, we design a continuous planning module. When S-Planner fails, the module stores the current environment locally. Upon invoking LLMs for replanning, it first loads the stored environment and incorporates it into the prompt, ensuring that LLMs continue planning from the failure step rather than starting over. This allows MeCo to seamlessly switch between two planning modules, significantly enhancing the collaborative efficiency between S-Planner and LLMs.

\subsection{Selective caching} \label{sec:D}
The number of tasks in the task cache affects the performance of the search process. As the number of tasks in the task cache increases, the time complexity of task matching also rises. Therefore, we set a maximum task count, denoted as $k$. When the task count reaches~$k$, tasks in the cache that are too similar or redundant will reduce the diversity of the tasks, preventing MeCo from efficiently leveraging the similarity between tasks for planning in diverse scenarios. In this case, a deduplication mechanism becomes crucial.
Inspired by the least frequently used (LFU) policy~\cite{willick1993disk}, we define $f(T_i)$ as the frequency with which task $T_i$ is used as a similar task. The task cache contains two tables to store task frequencies: one is a cache table for speeding up access, and the other is a global table for frequency updates. When planning a new task, MeCo first searches the cache table. If a similar task is found, $f(T_i)$ is incremented by 1, and the updated value of $f(T_i)$ is synchronized to the global table. If no similar task is found, the system invokes the LLMs for planning while asynchronously searching the global table and recording results for future deduplication. After completing a task, MeCo updates both the cache table and the global table based on task usage frequency, using the following principle:
\begin{equation}
C_{\text{cache}}' = \arg\max_{C \subseteq C_{\text{global}}} 
\; \sum_{T \in C} f(T), \quad \text{s.t. } |C| = k
\end{equation}
where $C_{\text{cache}}'$ denotes the updated cache table, 
$C_{\text{global}}$ represents the global table, 
and $k$ is the maximum task count in the cache.

The quality of tasks in the task cache also affects the performance of S-Planner. For example, in \textit{Pack Grocery} task, objects need to be placed inside a box. If an object is initially placed outside the box and then moved inside in a later step, the task is still considered completed. However, when such a task is referenced as a similar task, it introduces redundant steps and may even lead to planning failures. To address this issue, after a task is completed, we first check its effective step count \( S \). If \( S \leq S_{\text{base}} \), it passes the check, where \(S_{\text{base}}\) is the baseline step count.

\begin{figure*}[t]
    \centering
    \includegraphics[width=\linewidth]{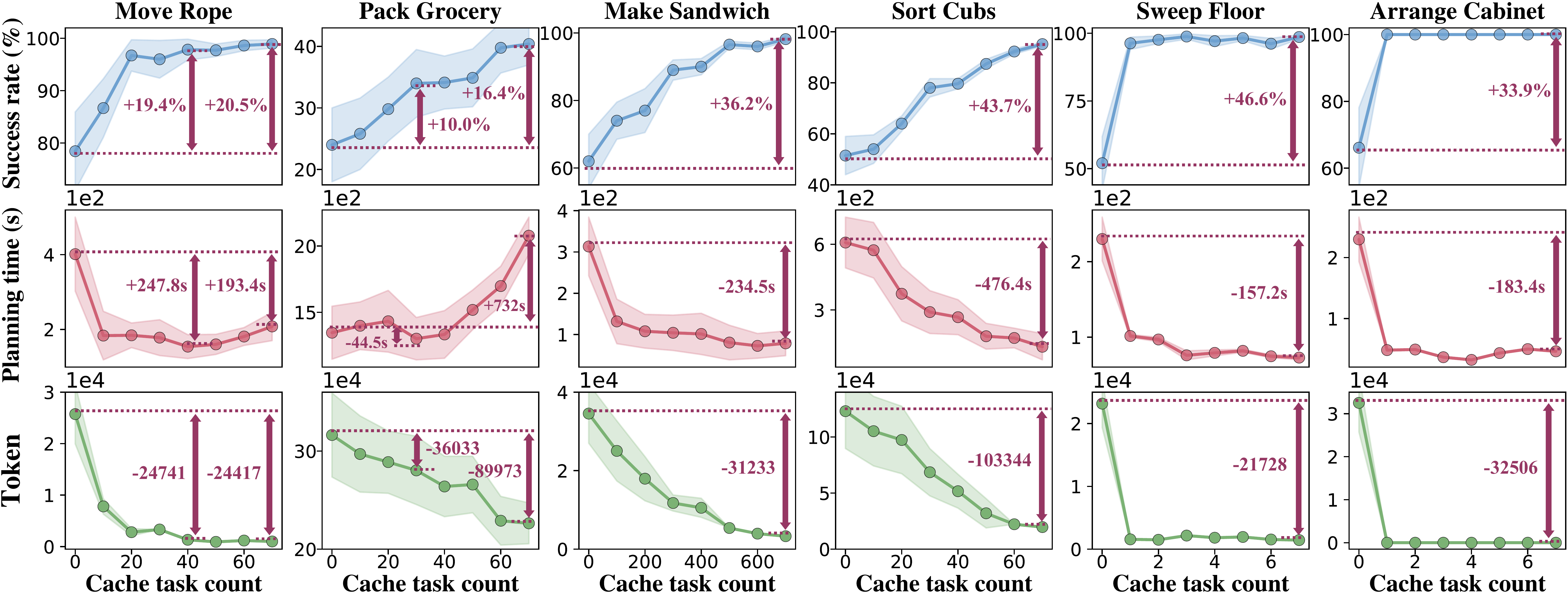}
    \caption{Performance vs. cache task count. As the number of tasks stored in the cache increases, we observe changes in the average success rate, planning time, and token consumption of MeCo across different tasks.  For each task type, we randomly store tasks planned by LLM-empowered methods in the task cache and evaluate MeCo averaged over 30 random seeds.}
    \label{k}
    \vspace{-0.1cm}
\end{figure*}

\section{Experimental evaluation} \label{sec:4}

We design a series of experiments to validate our methods. In Section \ref{sec:e1}, we first introduce the experimental setup. In Section \ref{sec:e2}, we analyze the impact of task counts stored in the cache on MeCo's performance and determine the optimal threshold for the maximum number of tasks in the cache. In Section \ref{sec:e3}, we evaluate MeCo's performance by comparing with four strong baselines. In Section \ref{sec:e4}, we conduct ablation studies on different components of MeCo, demonstrating the effectiveness of each design.

\subsection{Experimental setup}\label{sec:e1}

\textbf{Experimental Environment.} We evaluate the performance of our methods on a server with a 24-core Intel Core i9-14900HX~CPU, and we use DeepSeek-V3~\cite{liu2024deepseek} as the basic LLM policy for all experiments.
The simulation is based on MuJoCo~\cite{todorov2012mujoco} physics engine.

\textbf{MeCoBench.} We present a variant of RoCoBench~\cite{mandi2024roco} for similar tasks, called MeCoBench. In contrast to RoCoBench, which primarily focuses on each task in isolation, MeCoBench emphasizes inter-task similarity. For each task, we introduce similarity indicators to increase benchmark complexity. This allows MeCoBench to operate in three modes: randomly generating tasks that are similar to those in the task cache, different to them, or without considering similarity at all. 
More details can be found in the Appendix \ref{bench}.

\begin{figure*}[t]
    \centering
    \includegraphics[width=\linewidth]{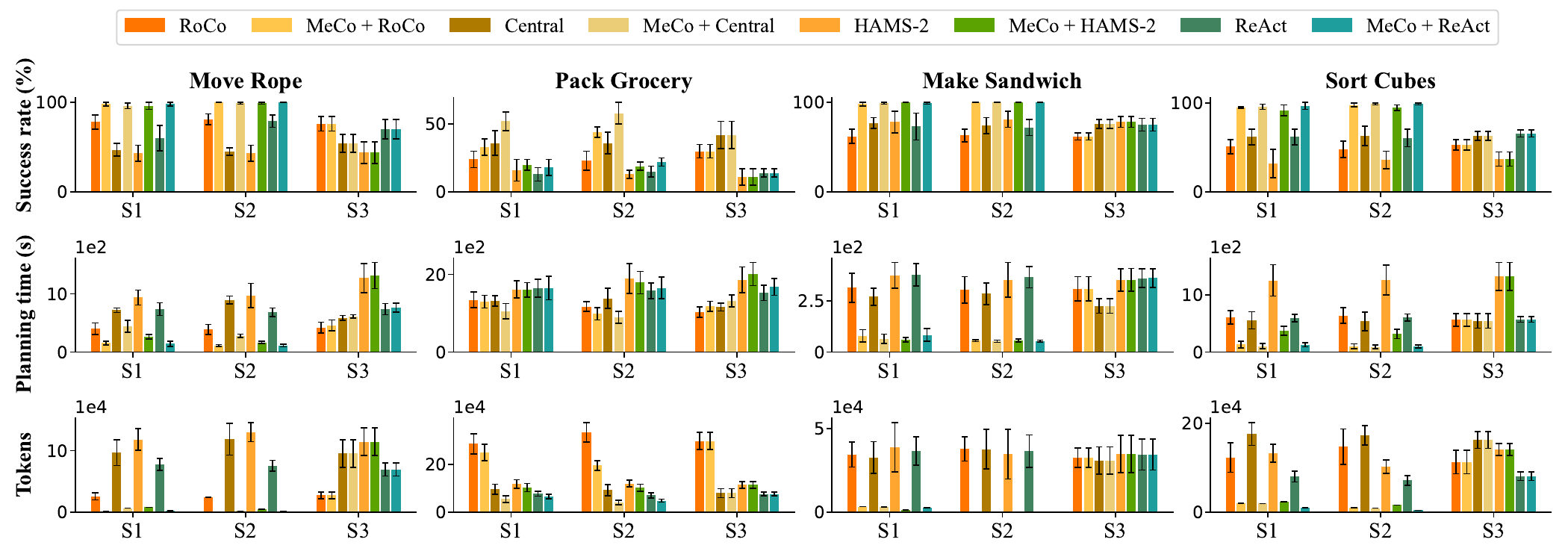}
    \caption{The performance of MeCo. We evaluate on MeCoBench across three scenarios: random (S1), totally similar (S2), and totally different (S3), using four baselines for comparison. For each task, we report success rate, planning time, and token consumption averaged over 30 random seeds. The results on \textit{Sweep Floor} and \textit{Arrange Cabinet} are provided in the Appendix \ref{A1}.}
    \label{simroco}
    \vspace{-0.2cm}
\end{figure*}

\textbf{Baseline Methods.} 
To evaluate the effectiveness of our framework, we compare it with the following baseline methods:
\begin{itemize}[left=1pt, itemsep=0pt, topsep=0pt, parsep=0pt]
    \item RoCo~\cite{mandi2024roco}: The baseline restricts each LLM agent to partial information and performs planning through dialogue. It achieves state-of-the-art performance in RoCoBench~\cite{mandi2024roco}.
    \item Central Plan: The baseline allows the LLM agent to access full environment observations and knowledge of all robots. It generates actions for all robots at once.
    \item HMAS-2~\cite{chen2024scalable}: The baseline first lets the central LLM planner generate a set of actions for each robot. Each robot is also equipped with an LLM agent that checks the assigned actions and provides feedback to the central planner, which then replans accordingly.
    \item ReAct~\cite{yao2023react}: The baseline enables the LLM agent to generate reasoning traces and task-specific actions in an interleaved manner.
\end{itemize}

\textbf{Evaluation Metric.} In a finite number of rounds with a limited number of attempts per round, we evaluate performance using the following metrics: (i) \textbf{Success rate}: the success rate of task planning, which assesses the algorithm's planning effectiveness. (ii) \textbf{Planning time}: the average time required from the start to the completion of planning, which measures planning efficiency. (iii) \textbf{Tokens}: The average token consumption, which reflects the algorithm's token usage. An algorithm performs better if it achieves a higher Success rate, lower Planning time, and lower Tokens. 
In addition, we use the standard deviation to measure the variability of success rates. However, once the cache reaches a certain number of tasks, the use of LLMs decreases. This leads to many small values and a few large values for \textbf{Planning time} and \textbf{Tokens}. In such cases, the standard deviation can even exceed the mean, making it unsuitable. To address this, we define a truncated standard deviation, 
which ignores outliers that deviate from the mean by more than $\tau$ times the standard deviation. This reduces the impact of extreme values on variability. In our experiments, we set $\tau=5$.

\subsection{Performance vs. cache task count}\label{sec:e2}
As the number of stored tasks in the task cache increases, task diversity also improves, enabling more efficient planning through task-level similarity in a wider range of scenarios. However, this also increases the time complexity during the search phase, leading to longer task-matching times. In this experiment, we study the impact of cache task count on MeCo's performance, as shown in Figure~\ref{k}. Meanwhile, to ensure higher success rates, shorter planning times, and lower token consumption, it is necessary to choose an appropriate maximum cache size for each task.

\textbf{High-workspace-overlap tasks} include \textit{Move Rope} and \textit{Pack Grocery}. As shown in the first two columns of Figure~\ref{k}, increasing the cache task count significantly improves success rates while substantially reducing token consumption. This indicates that a larger cache allows the system to better exploit task-level similarity across diverse scenarios, thereby reducing reliance on LLMs and improving planning efficiency. However, planning time follows a “decrease-then-increase” trend. With a small cache task count, the acceleration effect dominates and planning time drops significantly. As the cache continues to grow, the enlarged search space increases the cost of matching similar tasks, which gradually offsets and eventually surpasses the benefits of caching. For example, in \textit{Pack Grocery}, planning time rebounds once the cache task count exceeds 30, showing that an overly large cache imposes additional overhead.

\textbf{Low-workspace-overlap tasks} include \textit{Make Sandwich}, \textit{Sort Cubes}, \textit{Sweep Floor}, and \textit{Arrange Cabinet}. As shown in the last four columns of Figure~\ref{k}, success rates generally increase with larger cache task count but converge at different rates across tasks. For example, \textit{Make Sandwich} involves nearly 1,000 diverse scenarios, and its success rate approaches 100\% only when the cache task count reaches about 500. In contrast, \textit{Sort Cubes} stabilizes around a cache task count of 40, while \textit{Sweep Floor} and \textit{Arrange Cabinet}, which are inherently more similar, achieve near-perfect success rates when the cache task count is as small as 1. Unlike high-workspace-overlap tasks, the planning time for these tasks continues to decrease as the cache task count grows and eventually stabilizes, since the search overhead is negligible and does not cause a rebound. Meanwhile, token consumption consistently decreases with larger cache task count across all low-workspace-overlap tasks.

The results confirm the critical impact of cache task count on performance. For high-workspace-overlap tasks, an excessively large cache may increase search overhead, requiring a balance between leveraging task similarity and controlling search cost. For low-workspace-overlap tasks, however, a larger cache almost unilaterally improves performance, as the search overhead is negligible. Consequently, the optimal cache size varies across different task types. Moreover, MeCo allows the maximum cache size to be adjusted according to specific needs. A larger cache can reduce token consumption and improve success rates, but it may also result in longer planning times. Considering success rate, planning time, and token consumption together, we set the maximum cache sizes to 40, 30, 700, 50, 9, and 9 for the six tasks in subsequent experiments.

\begin{table*}[t]
\caption{Ablation study on the similarity testing, S-Planner, and continuous planning module. 
The experiments are conducted under the MeCo which uses RoCo~\cite{mandi2024roco} as the LLM-empowered method. Evaluation results are averaged over 30 random seeds per task.
Additional ablation studies on the selective caching mechanism are provided in the Appendix \ref{A2}.}
\label{rocosim_results}
\centering
\resizebox{\textwidth}{!}{
\begin{tblr}{
  colspec = {c c *{6}{c}}, 
  colsep = 4pt,
  cell{2-7}{2} = {font=\footnotesize},
  cell{2-7}{1} = {font=\small},
  cell{1}{3-8} = {font=\small},
  cell{2-7}{3-8} = {font=\small},
  hline{1} = {1pt},       
  hline{2} = {1pt},       
  hline{3-6} = {0.4pt},   
  hline{7} = {1pt},       
}
                               &                              & {Move \\ Rope} & {Pack \\ Grocery} & {Make \\ Sandwich} & {Sort \\ Cubes} & {Sweep \\ Floor} & {Arrange \\ Cabinet} \\
{MeCo w/o \\ Similarity testing}         & {Success rate \\ Time, Tokens} &  {0.84 $\pm$ 0.05 \textcolor{red}{$\downarrow$} \\ 447.81s \textcolor{red}{$\uparrow$}, 17465 \textcolor{red}{$\uparrow$}} & {0.25 $\pm$ 0.06 \textcolor{red}{$\downarrow$} \\ 1573.88s \textcolor{red}{$\uparrow$}, 281165 \textcolor{red}{$\uparrow$}} & {0.90 $\pm$ 0.04 \textcolor{red}{$\downarrow$} \\ 128.30s \textcolor{red}{$\uparrow$}, 7679 \textcolor{red}{$\uparrow$}} & {0.78 $\pm$ 0.09 \textcolor{red}{$\downarrow$} \\ 291.46s \textcolor{red}{$\uparrow$}, 68592 \textcolor{red}{$\uparrow$}} & {0.99 $\pm$ 0.10 \textcolor{green}{$\approx$} \\ 75.86s \textcolor{red}{$\uparrow$}, 2164 \textcolor{red}{$\uparrow$}} & {1.00 $\pm$ 0 \textcolor{green}{$\approx$} \\ 43.58s \textcolor{green}{$\approx$}, 0 \textcolor{green}{$\approx$}} \\
{MeCo w/o \\ S-Planner}      & {Success rate \\ Time, Tokens} &  {0.78 $\pm$ 0.08 \textcolor{red}{$\downarrow$} \\ 444.35s \textcolor{red}{$\uparrow$}, 25737 \textcolor{red}{$\uparrow$}} & {0.24 $\pm$ 0.06 \textcolor{red}{$\downarrow$} \\ 1502.32s \textcolor{red}{$\uparrow$}, 286297 \textcolor{red}{$\uparrow$}} & {0.62 $\pm$ 0.08 \textcolor{red}{$\downarrow$} \\ 313.38s \textcolor{red}{$\uparrow$}, 34537 \textcolor{red}{$\uparrow$}} & {0.51 $\pm$ 0.11 \textcolor{red}{$\downarrow$} \\ 608.80s \textcolor{red}{$\uparrow$}, 123028 \textcolor{red}{$\uparrow$}} & {0.52 $\pm$ 0.09 \textcolor{red}{$\downarrow$} \\ 229.98s \textcolor{red}{$\uparrow$}, 23171 \textcolor{red}{$\uparrow$}} & {0.66 $\pm$ 0.12 \textcolor{red}{$\downarrow$} \\ 229.79s \textcolor{red}{$\uparrow$}, 32505 \textcolor{red}{$\uparrow$}} \\
{MeCo w/o \\ Continuous Planning}    & {Success rate \\ Time, Tokens} &  {0.97 $\pm$ 0.01  \textcolor{red}{$\downarrow$} \\ 214.29s \textcolor{red}{$\uparrow$}, 2871 \textcolor{red}{$\uparrow$}} & {0.30 $\pm$ 0.04 \textcolor{red}{$\downarrow$} \\ 1381.33s \textcolor{red}{$\uparrow$}, 261346 \textcolor{red}{$\uparrow$}} & {0.98 $\pm$ 0.02 \textcolor{green}{$\approx$} \\ 81.15s \textcolor{red}{$\uparrow$}, 3368 \textcolor{red}{$\uparrow$}} & {0.91 $\pm$ 0.03 \textcolor{red}{$\downarrow$} \\ 180.37s \textcolor{red}{$\uparrow$}, 51671 \textcolor{red}{$\uparrow$}} & {0.99 $\pm$ 0.10 \textcolor{green}{$\approx$} \\ 79.11s \textcolor{red}{$\uparrow$}, 1821 \textcolor{red}{$\uparrow$}} & {1.00 $\pm$ 0 \textcolor{green}{$\approx$} \\ 45.94s \textcolor{green}{$\approx$}, 0 \textcolor{green}{$\approx$}} \\
\SetCell[r=2]{font=\bfseries\small}{MeCo} &
\SetCell[r=2]{font=\bfseries\footnotesize}{Success rate \\ Time, Tokens} &
\SetCell[r=2]{font=\bfseries\small}{0.98 ± 0.02 \\ 154.63s, 1320} &
\SetCell[r=2]{font=\bfseries\small}{0.33 ± 0.06 \\ 1300.98s, 250264} &
\SetCell[r=2]{font=\bfseries\small}{0.98 ± 0.02 \\ 78.78s, 3304} &
\SetCell[r=2]{font=\bfseries\small}{0.95 ± 0.01 \\ 132.70s, 19684} &
\SetCell[r=2]{font=\bfseries\small}{0.99 ± 0.20 \\ 72.82s, 1443} &
\SetCell[r=2]{font=\bfseries\small}{1.00 ± 0 \\ 47.82s, 0} \\

= & = & = & = & = & = & = & = \\ 
\end{tblr}
}
\end{table*}

\subsection{Performance of MeCo}\label{sec:e3}
We evaluate MeCo on MeCoBench under three scenarios: \textbf{random (S1)}, \textbf{totally similar (S2)}, and \textbf{totally different (S3)}. We use each of four baselines (RoCo~\cite{mandi2024roco}, Central Plan, HMAS-2~\cite{chen2024scalable}, and ReAct~\cite{yao2023react}) as the LLM-empowered module within MeCo, and compare it with these strong baselines. As shown in Figure~\ref{simroco}, MeCo significantly reduces token consumption while improving success rates and reducing planning time.

\textbf{Success rate.} The first row of Figure~\ref{simroco} presents the success rate performance of MeCo. In both the random (S1) and totally similar (S2) scenarios, MeCo achieves a significant improvement over the four baselines across all tasks. Specifically, the average success rate increases by about 29\% in the random scenario and 32\% in the similar scenario, indicating that MeCo performs better as task similarity increases. The \textit{Move Rope}, \textit{Make Sandwich}, and \textit{Sort Cubes} tasks nearly reach 100\%, suggesting that when latent task-level similarity exist, MeCo effectively captures and transfers useful knowledge, achieving near-optimal performance. However, for the \textit{Pack Grocery} task, all methods show relatively low success rates due to the high risk of collisions. Among them, ReAct and HMAS-2 reach only about 15\%. Even so, MeCo still improves performance by roughly 8\%, demonstrating its robustness in complex environments. Meanwhile, in the totally different (S3) scenario, MeCo performs on par with the baseline, indicating that it not only plans effectively under similar-task conditions but also maintains performance when tasks differ substantially.

\textbf{Planning time.} The second row of Figure~\ref{simroco} shows the reduction in planning time of MeCo. For the \textit{Move Rope}, \textit{Make Sandwich}, and \textit{Sort Cubes} tasks, MeCo significantly reduces planning time in both the random (S1) and totally similar (S2) scenarios. Specifically, in the random scenario, the average reductions are 64\% for \textit{Move Rope}, 78\% for \textit{Make Sandwich}, and 76\% for \textit{Sort Cubes}. This shows that MeCo effectively reduces LLMs inference time by leveraging task-level similarities to decrease the number of LLMs calls. In contrast, the reduction is only 5\% for \textit{Pack Grocery}, mainly because this task involves a time-consuming search process, and the time saved by referencing similar tasks is less significant. In the similar scenario, the average reductions reach 77\% for \textit{Move Rope}, 12\% for \textit{Pack Grocery}, 83\% for \textit{Make Sandwich}, and 80\% for \textit{Sort Cubes}, indicating that higher task similarity leads to greater time savings. However, in the totally different (S3) scenario, the planning time increases by roughly 6\%, suggesting that while MeCo introduces slight overhead in low-similarity conditions, the increase remains within an acceptable range.

\textbf{Token usage.} In the random (S1) scenario, MeCo significantly reduces token consumption compared with the baselines, achieving an average reduction of about 73\% across all tasks. For the \textit{Move Rope}, \textit{Make Sandwich}, and \textit{Sort Cubes} tasks, the reduction exceeds 90\%, demonstrating the effectiveness of MeCo in lowering planning costs. In the totally similar (S2) scenario, the average token reduction reaches around 82\%, as higher task similarity enables MeCo to make fewer LLM calls. However, \textit{Pack Grocery} shows only an 18\% reduction, mainly due to its inherently lower success rate. In the totally different (S3) scenario, token usage remains comparable to the baselines, indicating that MeCo does not introduce additional
overhead when task similarity is low.

\subsection{Ablation studies}\label{sec:e4}

MeCo consists of several key components, including the similarity testing in the search stage, the similar motion planner (S-Planner), the continuous planning module, and the selective caching mechanism. To investigate the role of each design, we conduct an ablation study.
Specifically, we select RoCo as the LLM-empowered method
for this experiment. First, we remove the similarity testing and instead select the task with the smallest average positional change of objects as the similar task for compensation. Second, we remove the S-Planner and directly reuse the previous trajectory as a substitute. Finally, we remove the continuous planning module, when S-Planner fails, the planning process is restarted from scratch.

As shown in Table~\ref{rocosim_results}, after removing the similarity testing, the success rate drops significantly. Especially in the \textit{Move Rope}, \textit{Pack Grocery}, \textit{Sort Cubes}, and \textit{Make Sandwich} tasks, the performance decreases by about 12\% on average. Meanwhile, both planning time and token consumption increase notably: \textit{Move Rope} by 189\%, \textit{Pack Grocery} by 21\%, \textit{Make Sandwich} by 63\%, and \textit{Sort Cubes} by 120\%. These results indicate that the absence of accurate similarity testing leads to ineffective task references and redundant reasoning, reducing overall efficiency. In contrast, for the \textit{Sweep Floor} and \textit{Arrange Cabinet} tasks, removing similarity testing causes no observable impact. This is because these tasks exhibit inherently high 
similarity, where even random task pairs share substantial resemblance, making complex filtering unnecessary. Overall, the results show that the similarity testing plays a crucial role in ensuring high-quality task search and preventing invalid matches.

Second, the performance degradation becomes more pronounced after removing the S-Planner. Since directly reusing previous trajectories results in nearly zero success rate, the overall success rate is almost the same as that of directly using LLM-based planning, showing an approximate 30\% drop compared with MeCo. Meanwhile, token consumption also aligns with that of direct LLM planning, exhibiting a significant increase. In particular, the planning time rises by 222.27\%, even exceeding that of direct LLM planning, as additional time is required for  matching similar tasks during the search stage. These results highlight the efficiency and superiority of the S-Planner in leveraging task-level similarity for planning.

Finally, after removing the continuous planning module, the success rate shows a slight decline in some tasks, while additional overhead is introduced, leading to a significant increase in both planning time and token consumption. This occurs because, without the continuous planning mechanism, LLMs must replan from scratch whenever the S-Planner fails locally, resulting in redundant reasoning and unnecessary computational cost. Moreover, the previous efforts of the S-Planner become wasted. Therefore, although this module has only a limited impact on the success rate, it plays a crucial role in improving overall reasoning efficiency.

\vspace{2.7pt}

\section{Conclusion} \label{sec:5}
In this paper, 
we propose MeCo, a similarity-aware multi-robot collaboration framework, that leverages task-level similarity for efficient planning.
In the search stage, we categorize tasks into high-workspace-overlap and low-workspace-overlap types based on robots’ workspace overlap levels. 
Furthermore, we design a novel similarity testing method to identify previously solved tasks with high relevance, enabling the reuse of existing plans without re-invoking the LLMs.
Within MeCo, the  similar motion planner (S-Planner) efficiently generates plans by leveraging prior task plans. When S-Planner fails, MeCo activates the continuous planning module, which resumes planning from the failed step using LLMs. Once a task is successfully planned, MeCo selectively caches it in the cache for future reuse. 
Experiments show that our approach significantly reduces planning costs while improving success rates.

\begin{acks}
This work was supported in part by the National Key R\&D Program of China (No. 2021YFB2900100), the National Science Fund for Distinguished Young Scholars (No. 62025205), and the National Natural Science Foundation of China (No. 62372383).
\end{acks}

\clearpage
\bibliographystyle{ACM-Reference-Format} 
\bibliography{ref}


\begin{thebibliography}{50}


\ifx \showCODEN    \undefined \def \showCODEN     #1{\unskip}     \fi
\ifx \showDOI      \undefined \def \showDOI       #1{#1}\fi
\ifx \showISBNx    \undefined \def \showISBNx     #1{\unskip}     \fi
\ifx \showISBNxiii \undefined \def \showISBNxiii  #1{\unskip}     \fi
\ifx \showISSN     \undefined \def \showISSN      #1{\unskip}     \fi
\ifx \showLCCN     \undefined \def \showLCCN      #1{\unskip}     \fi
\ifx \shownote     \undefined \def \shownote      #1{#1}          \fi
\ifx \showarticletitle \undefined \def \showarticletitle #1{#1}   \fi
\ifx \showURL      \undefined \def \showURL       {\relax}        \fi
\providecommand\bibfield[2]{#2}
\providecommand\bibinfo[2]{#2}
\providecommand\natexlab[1]{#1}
\providecommand\showeprint[2][]{arXiv:#2}

\bibitem[\protect\citeauthoryear{??}{ama}{[n.d.]a}]%
        {amazonUniversalModel}

\newblock \bibinfo{title}{{H}ow a universal model is helping one generation of
  {A}mazon robots train the next}.
\newblock
  \bibinfo{howpublished}{\url{https://www.amazon.science/latest-news/how-a-universal-model-is-helping-one-generation-of-amazon-robots-train-the-next}}.
\newblock
\newblock
\shownote{[Accessed 04-02-2025].}


\bibitem[\protect\citeauthoryear{??}{ama}{[n.d.]b}]%
        {amazonPinchgraspingRobot}

\newblock \bibinfo{title}{{P}inch-grasping robot handles items with precision}.
\newblock
  \bibinfo{howpublished}{\url{https://www.amazon.science/latest-news/pinch-grasping-robot-handles-items-with-precision}}.
\newblock
\newblock
\shownote{[Accessed 04-02-2025].}


\bibitem[\protect\citeauthoryear{??}{vis}{[n.d.]}]%
        {visualcomponentsRobotOffline}

\newblock \bibinfo{title}{{R}obot offline programming}.
\newblock
  \bibinfo{howpublished}{\url{https://www.visualcomponents.com/products/robot-offline-programming/}}.
\newblock
\newblock
\shownote{[Accessed 04-02-2025].}


\bibitem[\protect\citeauthoryear{Agrawal, Nayak, Naik, and Hayes}{Agrawal
  et~al\mbox{.}}{2023}]%
        {agrawal2023shelfhelp}
\bibfield{author}{\bibinfo{person}{Shivendra Agrawal}, \bibinfo{person}{Suresh
  Nayak}, \bibinfo{person}{Ashutosh Naik}, {and} \bibinfo{person}{Bradley
  Hayes}.} \bibinfo{year}{2023}\natexlab{}.
\newblock \showarticletitle{ShelfHelp: Empowering Humans to Perform
  Vision-Independent Manipulation Tasks with a Socially Assistive Robotic
  Cane}. In \bibinfo{booktitle}{\emph{Proc. of AAMAS}}.
\newblock


\bibitem[\protect\citeauthoryear{Ahn, Brohan, Brown, Chebotar, Cortes, David,
  Finn, Fu, Gopalakrishnan, Hausman, et~al\mbox{.}}{Ahn et~al\mbox{.}}{2022}]%
        {ahn2022can}
\bibfield{author}{\bibinfo{person}{Michael Ahn}, \bibinfo{person}{Anthony
  Brohan}, \bibinfo{person}{Noah Brown}, \bibinfo{person}{Yevgen Chebotar},
  \bibinfo{person}{Omar Cortes}, \bibinfo{person}{Byron David},
  \bibinfo{person}{Chelsea Finn}, \bibinfo{person}{Chuyuan Fu},
  \bibinfo{person}{Keerthana Gopalakrishnan}, \bibinfo{person}{Karol Hausman},
  {et~al\mbox{.}}} \bibinfo{year}{2022}\natexlab{}.
\newblock \showarticletitle{Do as {I} can, not as {I} say: Grounding language
  in robotic affordances}.
\newblock \bibinfo{journal}{\emph{arXiv preprint arXiv:2204.01691}}
  (\bibinfo{year}{2022}).
\newblock


\bibitem[\protect\citeauthoryear{Brown, Mann, Ryder, Subbiah, Kaplan, Dhariwal,
  Neelakantan, Shyam, Sastry, Askell, et~al\mbox{.}}{Brown
  et~al\mbox{.}}{2020}]%
        {brown2020language}
\bibfield{author}{\bibinfo{person}{Tom Brown}, \bibinfo{person}{Benjamin Mann},
  \bibinfo{person}{Nick Ryder}, \bibinfo{person}{Melanie Subbiah},
  \bibinfo{person}{Jared~D Kaplan}, \bibinfo{person}{Prafulla Dhariwal},
  \bibinfo{person}{Arvind Neelakantan}, \bibinfo{person}{Pranav Shyam},
  \bibinfo{person}{Girish Sastry}, \bibinfo{person}{Amanda Askell},
  {et~al\mbox{.}}} \bibinfo{year}{2020}\natexlab{}.
\newblock \showarticletitle{Language models are few-shot learners}. In
  \bibinfo{booktitle}{\emph{Proc. of NeurIPS}}.
\newblock


\bibitem[\protect\citeauthoryear{Chen, Lin, Liu, Ma, Liang, and Wong}{Chen
  et~al\mbox{.}}{2025b}]%
        {chen2025affordances}
\bibfield{author}{\bibinfo{person}{Jiaqi Chen}, \bibinfo{person}{Bingqian Lin},
  \bibinfo{person}{Xinmin Liu}, \bibinfo{person}{Lin Ma},
  \bibinfo{person}{Xiaodan Liang}, {and} \bibinfo{person}{Kwan-Yee~K Wong}.}
  \bibinfo{year}{2025}\natexlab{b}.
\newblock \showarticletitle{Affordances-oriented planning using foundation
  models for continuous vision-language navigation}. In
  \bibinfo{booktitle}{\emph{Proc. of AAAI}}.
\newblock


\bibitem[\protect\citeauthoryear{Chen, Yu, Zhou, Xu, Mu, Hu, Shao, Wang, Li,
  and Shao}{Chen et~al\mbox{.}}{2025c}]%
        {chen2024emos}
\bibfield{author}{\bibinfo{person}{Junting Chen}, \bibinfo{person}{Checheng
  Yu}, \bibinfo{person}{Xunzhe Zhou}, \bibinfo{person}{Tianqi Xu},
  \bibinfo{person}{Yao Mu}, \bibinfo{person}{Mengkang Hu},
  \bibinfo{person}{Wenqi Shao}, \bibinfo{person}{Yikai Wang},
  \bibinfo{person}{Guohao Li}, {and} \bibinfo{person}{Lin Shao}.}
  \bibinfo{year}{2025}\natexlab{c}.
\newblock \showarticletitle{EMOS: Embodiment-aware Heterogeneous Multi-robot
  Operating System with LLM Agents}. In \bibinfo{booktitle}{\emph{Proc. of
  ICLR}}.
\newblock


\bibitem[\protect\citeauthoryear{Chen, Guo, Wang, Li, Zhao, Liu, Ding, Pan, and
  Yu}{Chen et~al\mbox{.}}{2025a}]%
        {chen2025future}
\bibfield{author}{\bibinfo{person}{Mengqi Chen}, \bibinfo{person}{Bin Guo},
  \bibinfo{person}{Hao Wang}, \bibinfo{person}{Haoyu Li}, \bibinfo{person}{Qian
  Zhao}, \bibinfo{person}{Jingqi Liu}, \bibinfo{person}{Yasan Ding},
  \bibinfo{person}{Yan Pan}, {and} \bibinfo{person}{Zhiwen Yu}.}
  \bibinfo{year}{2025}\natexlab{a}.
\newblock \showarticletitle{The future of cognitive strategy-enhanced
  persuasive dialogue agents: new perspectives and trends}.
\newblock \bibinfo{journal}{\emph{Frontiers of Computer Science}}
  (\bibinfo{year}{2025}).
\newblock


\bibitem[\protect\citeauthoryear{Chen, You, Li, Guan, Qian, Zhao, Yang, Xie,
  Liu, and Sun}{Chen et~al\mbox{.}}{2024b}]%
        {chen2024internet}
\bibfield{author}{\bibinfo{person}{Weize Chen}, \bibinfo{person}{Ziming You},
  \bibinfo{person}{Ran Li}, \bibinfo{person}{Yitong Guan},
  \bibinfo{person}{Chen Qian}, \bibinfo{person}{Chenyang Zhao},
  \bibinfo{person}{Cheng Yang}, \bibinfo{person}{Ruobing Xie},
  \bibinfo{person}{Zhiyuan Liu}, {and} \bibinfo{person}{Maosong Sun}.}
  \bibinfo{year}{2024}\natexlab{b}.
\newblock \showarticletitle{Internet of agents: Weaving a web of heterogeneous
  agents for collaborative intelligence}.
\newblock \bibinfo{journal}{\emph{arXiv preprint arXiv:2407.07061}}
  (\bibinfo{year}{2024}).
\newblock


\bibitem[\protect\citeauthoryear{Chen, Arkin, Zhang, Roy, and Fan}{Chen
  et~al\mbox{.}}{2024a}]%
        {chen2024scalable}
\bibfield{author}{\bibinfo{person}{Yongchao Chen}, \bibinfo{person}{Jacob
  Arkin}, \bibinfo{person}{Yang Zhang}, \bibinfo{person}{Nicholas Roy}, {and}
  \bibinfo{person}{Chuchu Fan}.} \bibinfo{year}{2024}\natexlab{a}.
\newblock \showarticletitle{Scalable multi-robot collaboration with large
  language models: Centralized or decentralized systems?}. In
  \bibinfo{booktitle}{\emph{Proc. of IEEE ICRA}}.
\newblock


\bibitem[\protect\citeauthoryear{Cui, Xu, Zhong, Xu, Shen, and Tang}{Cui
  et~al\mbox{.}}{2024}]%
        {cui2024task}
\bibfield{author}{\bibinfo{person}{Yuanzhe Cui}, \bibinfo{person}{Zhipeng Xu},
  \bibinfo{person}{Lou Zhong}, \bibinfo{person}{Pengjie Xu},
  \bibinfo{person}{Yichao Shen}, {and} \bibinfo{person}{Qirong Tang}.}
  \bibinfo{year}{2024}\natexlab{}.
\newblock \showarticletitle{A Task-Adaptive Deep Reinforcement Learning
  Framework for Dual-Arm Robot Manipulation}.
\newblock \bibinfo{journal}{\emph{IEEE TASAE}} (\bibinfo{year}{2024}).
\newblock


\bibitem[\protect\citeauthoryear{Goldberg}{Goldberg}{2019}]%
        {goldberg2019robots}
\bibfield{author}{\bibinfo{person}{Ken Goldberg}.}
  \bibinfo{year}{2019}\natexlab{}.
\newblock \showarticletitle{Robots and the return to collaborative
  intelligence}.
\newblock \bibinfo{journal}{\emph{Nature Machine Intelligence}}
  (\bibinfo{year}{2019}).
\newblock


\bibitem[\protect\citeauthoryear{Ho and Ermon}{Ho and Ermon}{2016}]%
        {ho2016generative}
\bibfield{author}{\bibinfo{person}{Jonathan Ho} {and} \bibinfo{person}{Stefano
  Ermon}.} \bibinfo{year}{2016}\natexlab{}.
\newblock \showarticletitle{Generative adversarial imitation learning}. In
  \bibinfo{booktitle}{\emph{Proc. of NeurIPS}}.
\newblock


\bibitem[\protect\citeauthoryear{Huang, Xia, Xiao, Chan, Liang, Florence, Zeng,
  Tompson, Mordatch, Chebotar, et~al\mbox{.}}{Huang et~al\mbox{.}}{2023}]%
        {huang2023inner}
\bibfield{author}{\bibinfo{person}{Wenlong Huang}, \bibinfo{person}{Fei Xia},
  \bibinfo{person}{Ted Xiao}, \bibinfo{person}{Harris Chan},
  \bibinfo{person}{Jacky Liang}, \bibinfo{person}{Pete Florence},
  \bibinfo{person}{Andy Zeng}, \bibinfo{person}{Jonathan Tompson},
  \bibinfo{person}{Igor Mordatch}, \bibinfo{person}{Yevgen Chebotar},
  {et~al\mbox{.}}} \bibinfo{year}{2023}\natexlab{}.
\newblock \showarticletitle{Inner Monologue: Embodied Reasoning through
  Planning with Language Models}. In \bibinfo{booktitle}{\emph{Proc. of CoRL}}.
\newblock


\bibitem[\protect\citeauthoryear{Ji, Chen, Zhang, Kompella, Fan, Liu, and
  Chang}{Ji et~al\mbox{.}}{2025}]%
        {ji2025collision}
\bibfield{author}{\bibinfo{person}{Jiabao Ji}, \bibinfo{person}{Yongchao Chen},
  \bibinfo{person}{Yang Zhang}, \bibinfo{person}{Ramana~Rao Kompella},
  \bibinfo{person}{Chuchu Fan}, \bibinfo{person}{Gaowen Liu}, {and}
  \bibinfo{person}{Shiyu Chang}.} \bibinfo{year}{2025}\natexlab{}.
\newblock \showarticletitle{Collision-and Reachability-Aware Multi-Robot
  Control with Grounded LLM Planners}.
\newblock \bibinfo{journal}{\emph{arXiv preprint arXiv:2505.20573}}
  (\bibinfo{year}{2025}).
\newblock


\bibitem[\protect\citeauthoryear{Ju, Juan, Gomez, Nakamura, and Li}{Ju
  et~al\mbox{.}}{2022}]%
        {ju2022transferring}
\bibfield{author}{\bibinfo{person}{Hao Ju}, \bibinfo{person}{Rongshun Juan},
  \bibinfo{person}{Randy Gomez}, \bibinfo{person}{Keisuke Nakamura}, {and}
  \bibinfo{person}{Guangliang Li}.} \bibinfo{year}{2022}\natexlab{}.
\newblock \showarticletitle{Transferring policy of deep reinforcement learning
  from simulation to reality for robotics}.
\newblock \bibinfo{journal}{\emph{Nature Machine Intelligence}}
  (\bibinfo{year}{2022}).
\newblock


\bibitem[\protect\citeauthoryear{Lai, Go, Li, Kr{\"o}ger, Schaal, Allen, and
  Scholz}{Lai et~al\mbox{.}}{2025}]%
        {lai2025roboballet}
\bibfield{author}{\bibinfo{person}{Matthew Lai}, \bibinfo{person}{Keegan Go},
  \bibinfo{person}{Zhibin Li}, \bibinfo{person}{Torsten Kr{\"o}ger},
  \bibinfo{person}{Stefan Schaal}, \bibinfo{person}{Kelsey Allen}, {and}
  \bibinfo{person}{Jonathan Scholz}.} \bibinfo{year}{2025}\natexlab{}.
\newblock \showarticletitle{RoboBallet: Planning for multirobot reaching with
  graph neural networks and reinforcement learning}.
\newblock \bibinfo{journal}{\emph{Science Robotics}} (\bibinfo{year}{2025}).
\newblock


\bibitem[\protect\citeauthoryear{Lee, Baker, Bederson, and Rapoport}{Lee
  et~al\mbox{.}}{2024}]%
        {lee2024levels}
\bibfield{author}{\bibinfo{person}{Audrey Lee}, \bibinfo{person}{Turner~S
  Baker}, \bibinfo{person}{Joshua~B Bederson}, {and}
  \bibinfo{person}{Benjamin~I Rapoport}.} \bibinfo{year}{2024}\natexlab{}.
\newblock \showarticletitle{{Levels of autonomy in FDA-cleared surgical robots:
  a systematic review}}.
\newblock \bibinfo{journal}{\emph{NPJ Digital Medicine}}
  (\bibinfo{year}{2024}).
\newblock


\bibitem[\protect\citeauthoryear{Leidner, Bejjani, Albu-Sch{\"a}ffer, and
  Beetz}{Leidner et~al\mbox{.}}{2016}]%
        {leidner2016robotic}
\bibfield{author}{\bibinfo{person}{Daniel Leidner}, \bibinfo{person}{Wissam
  Bejjani}, \bibinfo{person}{Alin~Olimpiu Albu-Sch{\"a}ffer}, {and}
  \bibinfo{person}{Michael Beetz}.} \bibinfo{year}{2016}\natexlab{}.
\newblock \showarticletitle{Robotic agents representing, reasoning, and
  executing wiping tasks for daily household chores}. In
  \bibinfo{booktitle}{\emph{Proc. of AAMAS}}.
\newblock


\bibitem[\protect\citeauthoryear{Liang, Huang, Xia, Xu, Hausman, Ichter,
  Florence, and Zeng}{Liang et~al\mbox{.}}{2023}]%
        {liang2023code}
\bibfield{author}{\bibinfo{person}{Jacky Liang}, \bibinfo{person}{Wenlong
  Huang}, \bibinfo{person}{Fei Xia}, \bibinfo{person}{Peng Xu},
  \bibinfo{person}{Karol Hausman}, \bibinfo{person}{Brian Ichter},
  \bibinfo{person}{Pete Florence}, {and} \bibinfo{person}{Andy Zeng}.}
  \bibinfo{year}{2023}\natexlab{}.
\newblock \showarticletitle{Code as policies: Language model programs for
  embodied control}. In \bibinfo{booktitle}{\emph{Proc. of IEEE ICRA}}.
\newblock


\bibitem[\protect\citeauthoryear{Lin, Wei-Kocsis, Zhang, Min, Gan, Asunda, and
  Athinarayanan}{Lin et~al\mbox{.}}{2025}]%
        {lin2025think}
\bibfield{author}{\bibinfo{person}{Wenjie Lin}, \bibinfo{person}{Jin
  Wei-Kocsis}, \bibinfo{person}{Jiansong Zhang}, \bibinfo{person}{Byung-Cheol
  Min}, \bibinfo{person}{Dongming Gan}, \bibinfo{person}{Paul Asunda}, {and}
  \bibinfo{person}{Ragu Athinarayanan}.} \bibinfo{year}{2025}\natexlab{}.
\newblock \showarticletitle{Think, Reflect, Create: Metacognitive Learning for
  Zero-Shot Robotic Planning with LLMs}.
\newblock \bibinfo{journal}{\emph{arXiv preprint arXiv:2505.14899}}
  (\bibinfo{year}{2025}).
\newblock


\bibitem[\protect\citeauthoryear{Liu, Feng, Xue, Wang, Wu, Lu, Zhao, Deng,
  Zhang, Ruan, et~al\mbox{.}}{Liu et~al\mbox{.}}{2024}]%
        {liu2024deepseek}
\bibfield{author}{\bibinfo{person}{Aixin Liu}, \bibinfo{person}{Bei Feng},
  \bibinfo{person}{Bing Xue}, \bibinfo{person}{Bingxuan Wang},
  \bibinfo{person}{Bochao Wu}, \bibinfo{person}{Chengda Lu},
  \bibinfo{person}{Chenggang Zhao}, \bibinfo{person}{Chengqi Deng},
  \bibinfo{person}{Chenyu Zhang}, \bibinfo{person}{Chong Ruan},
  {et~al\mbox{.}}} \bibinfo{year}{2024}\natexlab{}.
\newblock \showarticletitle{Deepseek-v3 technical report}.
\newblock \bibinfo{journal}{\emph{arXiv preprint arXiv:2412.19437}}
  (\bibinfo{year}{2024}).
\newblock


\bibitem[\protect\citeauthoryear{Mahler, Matl, Satish, Danielczuk, DeRose,
  McKinley, and Goldberg}{Mahler et~al\mbox{.}}{2019}]%
        {mahler2019learning}
\bibfield{author}{\bibinfo{person}{Jeffrey Mahler}, \bibinfo{person}{Matthew
  Matl}, \bibinfo{person}{Vishal Satish}, \bibinfo{person}{Michael Danielczuk},
  \bibinfo{person}{Bill DeRose}, \bibinfo{person}{Stephen McKinley}, {and}
  \bibinfo{person}{Ken Goldberg}.} \bibinfo{year}{2019}\natexlab{}.
\newblock \showarticletitle{Learning ambidextrous robot grasping policies}.
\newblock \bibinfo{journal}{\emph{Science Robotics}} (\bibinfo{year}{2019}).
\newblock


\bibitem[\protect\citeauthoryear{Mandi, Jain, and Song}{Mandi
  et~al\mbox{.}}{2024}]%
        {mandi2024roco}
\bibfield{author}{\bibinfo{person}{Zhao Mandi}, \bibinfo{person}{Shreeya Jain},
  {and} \bibinfo{person}{Shuran Song}.} \bibinfo{year}{2024}\natexlab{}.
\newblock \showarticletitle{Roco: Dialectic multi-robot collaboration with
  large language models}. In \bibinfo{booktitle}{\emph{Proc. of IEEE ICRA}}.
\newblock


\bibitem[\protect\citeauthoryear{Mao, Liao, Yuan, and Zhu}{Mao
  et~al\mbox{.}}{2024}]%
        {mao2024multimodal}
\bibfield{author}{\bibinfo{person}{Qian Mao}, \bibinfo{person}{Zijian Liao},
  \bibinfo{person}{Jinfeng Yuan}, {and} \bibinfo{person}{Rong Zhu}.}
  \bibinfo{year}{2024}\natexlab{}.
\newblock \showarticletitle{Multimodal tactile sensing fused with vision for
  dexterous robotic housekeeping}.
\newblock \bibinfo{journal}{\emph{Nature Communications}}
  (\bibinfo{year}{2024}).
\newblock


\bibitem[\protect\citeauthoryear{Mehta, Bimbraw, Chittawadigi, and Saha}{Mehta
  et~al\mbox{.}}{2016}]%
        {mehta2016teach}
\bibfield{author}{\bibinfo{person}{Ishaan Mehta}, \bibinfo{person}{Keshav
  Bimbraw}, \bibinfo{person}{Rajeevlochana~G Chittawadigi}, {and}
  \bibinfo{person}{Subir~K Saha}.} \bibinfo{year}{2016}\natexlab{}.
\newblock \showarticletitle{{A teach pendant to control virtual robots in
  Roboanalyzer}}. In \bibinfo{booktitle}{\emph{IEEE International Conference on
  Robotics and Automation for Humanitarian Applications}}.
\newblock


\bibitem[\protect\citeauthoryear{Mon-Williams, Li, Long, Du, and
  Lucas}{Mon-Williams et~al\mbox{.}}{2025}]%
        {mon2025embodied}
\bibfield{author}{\bibinfo{person}{Ruaridh Mon-Williams}, \bibinfo{person}{Gen
  Li}, \bibinfo{person}{Ran Long}, \bibinfo{person}{Wenqian Du}, {and}
  \bibinfo{person}{Christopher~G Lucas}.} \bibinfo{year}{2025}\natexlab{}.
\newblock \showarticletitle{Embodied large language models enable robots to
  complete complex tasks in unpredictable environments}.
\newblock \bibinfo{journal}{\emph{Nature Machine Intelligence}}
  (\bibinfo{year}{2025}).
\newblock


\bibitem[\protect\citeauthoryear{Pal, Chauhan, and Baranwal}{Pal
  et~al\mbox{.}}{2025}]%
        {pal2025together}
\bibfield{author}{\bibinfo{person}{Aritra Pal}, \bibinfo{person}{Anandsingh
  Chauhan}, {and} \bibinfo{person}{Mayank Baranwal}.}
  \bibinfo{year}{2025}\natexlab{}.
\newblock \showarticletitle{Together We Rise: Optimizing Real-Time Multi-Robot
  Task Allocation using Coordinated Heterogeneous Plays}. In
  \bibinfo{booktitle}{\emph{Proc. of AAMAS}}.
\newblock


\bibitem[\protect\citeauthoryear{Radford, Wu, Child, Luan, Amodei, Sutskever,
  et~al\mbox{.}}{Radford et~al\mbox{.}}{2019}]%
        {radford2019language}
\bibfield{author}{\bibinfo{person}{Alec Radford}, \bibinfo{person}{Jeffrey Wu},
  \bibinfo{person}{Rewon Child}, \bibinfo{person}{David Luan},
  \bibinfo{person}{Dario Amodei}, \bibinfo{person}{Ilya Sutskever},
  {et~al\mbox{.}}} \bibinfo{year}{2019}\natexlab{}.
\newblock \showarticletitle{Language models are unsupervised multitask
  learners}.
\newblock \bibinfo{journal}{\emph{OpenAI blog}} (\bibinfo{year}{2019}).
\newblock


\bibitem[\protect\citeauthoryear{Raffel, Shazeer, Roberts, Lee, Narang, Matena,
  Zhou, Li, and Liu}{Raffel et~al\mbox{.}}{2020}]%
        {raffel2020exploring}
\bibfield{author}{\bibinfo{person}{Colin Raffel}, \bibinfo{person}{Noam
  Shazeer}, \bibinfo{person}{Adam Roberts}, \bibinfo{person}{Katherine Lee},
  \bibinfo{person}{Sharan Narang}, \bibinfo{person}{Michael Matena},
  \bibinfo{person}{Yanqi Zhou}, \bibinfo{person}{Wei Li}, {and}
  \bibinfo{person}{Peter~J Liu}.} \bibinfo{year}{2020}\natexlab{}.
\newblock \showarticletitle{Exploring the limits of transfer learning with a
  unified text-to-text transformer}.
\newblock \bibinfo{journal}{\emph{Journal of machine learning research}}
  (\bibinfo{year}{2020}).
\newblock


\bibitem[\protect\citeauthoryear{Schmidgall, Kim, Kuntz, Ghazi, and
  Krieger}{Schmidgall et~al\mbox{.}}{2024}]%
        {schmidgall2024general}
\bibfield{author}{\bibinfo{person}{Samuel Schmidgall},
  \bibinfo{person}{Ji~Woong Kim}, \bibinfo{person}{Alan Kuntz},
  \bibinfo{person}{Ahmed~Ezzat Ghazi}, {and} \bibinfo{person}{Axel Krieger}.}
  \bibinfo{year}{2024}\natexlab{}.
\newblock \showarticletitle{General-purpose foundation models for increased
  autonomy in robot-assisted surgery}.
\newblock \bibinfo{journal}{\emph{Nature Machine Intelligence}}
  (\bibinfo{year}{2024}).
\newblock


\bibitem[\protect\citeauthoryear{Seo and Unhelkar}{Seo and Unhelkar}{2025}]%
        {seo2025hierarchical}
\bibfield{author}{\bibinfo{person}{Sangwon Seo} {and} \bibinfo{person}{Vaibhav
  Unhelkar}.} \bibinfo{year}{2025}\natexlab{}.
\newblock \showarticletitle{Hierarchical Imitation Learning of Team Behavior
  from Heterogeneous Demonstrations}. In \bibinfo{booktitle}{\emph{Proc. of
  AAMAS}}.
\newblock


\bibitem[\protect\citeauthoryear{Singh, Blukis, Mousavian, Goyal, Xu, Tremblay,
  Fox, Thomason, and Garg}{Singh et~al\mbox{.}}{2023}]%
        {singh2023progprompt}
\bibfield{author}{\bibinfo{person}{Ishika Singh}, \bibinfo{person}{Valts
  Blukis}, \bibinfo{person}{Arsalan Mousavian}, \bibinfo{person}{Ankit Goyal},
  \bibinfo{person}{Danfei Xu}, \bibinfo{person}{Jonathan Tremblay},
  \bibinfo{person}{Dieter Fox}, \bibinfo{person}{Jesse Thomason}, {and}
  \bibinfo{person}{Animesh Garg}.} \bibinfo{year}{2023}\natexlab{}.
\newblock \showarticletitle{Progprompt: Generating situated robot task plans
  using large language models}. In \bibinfo{booktitle}{\emph{Proc. of IEEE
  ICRA}}.
\newblock


\bibitem[\protect\citeauthoryear{Todorov, Erez, and Tassa}{Todorov
  et~al\mbox{.}}{2012}]%
        {todorov2012mujoco}
\bibfield{author}{\bibinfo{person}{Emanuel Todorov}, \bibinfo{person}{Tom
  Erez}, {and} \bibinfo{person}{Yuval Tassa}.} \bibinfo{year}{2012}\natexlab{}.
\newblock \showarticletitle{Mujoco: A physics engine for model-based control}.
  In \bibinfo{booktitle}{\emph{Proc. of IEEE IROS}}.
\newblock


\bibitem[\protect\citeauthoryear{Wan, Du, Ibrahim, Qian, Jabbour, Zhao,
  Krishna, Raychowdhury, and Reddi}{Wan et~al\mbox{.}}{2025}]%
        {wan2025reca}
\bibfield{author}{\bibinfo{person}{Zishen Wan}, \bibinfo{person}{Yuhang Du},
  \bibinfo{person}{Mohamed Ibrahim}, \bibinfo{person}{Jiayi Qian},
  \bibinfo{person}{Jason Jabbour}, \bibinfo{person}{Yang Zhao},
  \bibinfo{person}{Tushar Krishna}, \bibinfo{person}{Arijit Raychowdhury},
  {and} \bibinfo{person}{Vijay~Janapa Reddi}.} \bibinfo{year}{2025}\natexlab{}.
\newblock \showarticletitle{ReCA: Integrated Acceleration for Real-Time and
  Efficient Cooperative Embodied Autonomous Agents}. In
  \bibinfo{booktitle}{\emph{Proc. of ACM ASPLOS}}.
\newblock


\bibitem[\protect\citeauthoryear{Wang, Yu, Zhang, Liu, Zeng, Zhou, Guo, and
  Xing}{Wang et~al\mbox{.}}{2025}]%
        {wang2025both}
\bibfield{author}{\bibinfo{person}{Hui Wang}, \bibinfo{person}{Zhiwen Yu},
  \bibinfo{person}{Yao Zhang}, \bibinfo{person}{Jiaqi Liu},
  \bibinfo{person}{Liekang Zeng}, \bibinfo{person}{Huan Zhou},
  \bibinfo{person}{Bin Guo}, {and} \bibinfo{person}{Guoliang Xing}.}
  \bibinfo{year}{2025}\natexlab{}.
\newblock \showarticletitle{BOTH: Efficient Coordination of Mobile Agents With
  Graph-Enhanced Bayesian Online Learning}.
\newblock \bibinfo{journal}{\emph{IEEE TMC}} (\bibinfo{year}{2025}).
\newblock


\bibitem[\protect\citeauthoryear{Wang, Yu, Zhang, Wang, Yang, Wang, Liu, and
  Guo}{Wang et~al\mbox{.}}{2024b}]%
        {wang2024hmos}
\bibfield{author}{\bibinfo{person}{Hui Wang}, \bibinfo{person}{Zhiwen Yu},
  \bibinfo{person}{Yao Zhang}, \bibinfo{person}{Yanfei Wang},
  \bibinfo{person}{Fan Yang}, \bibinfo{person}{Liang Wang},
  \bibinfo{person}{Jiaqi Liu}, {and} \bibinfo{person}{Bin Guo}.}
  \bibinfo{year}{2024}\natexlab{b}.
\newblock \showarticletitle{hmos: An extensible platform for task-oriented
  human--machine computing}.
\newblock \bibinfo{journal}{\emph{IEEE Transactions on Human-Machine Systems}}
  (\bibinfo{year}{2024}).
\newblock


\bibitem[\protect\citeauthoryear{Wang, He, and Kantaros}{Wang
  et~al\mbox{.}}{2024a}]%
        {wang2024probabilistically}
\bibfield{author}{\bibinfo{person}{Jun Wang}, \bibinfo{person}{Guocheng He},
  {and} \bibinfo{person}{Yiannis Kantaros}.} \bibinfo{year}{2024}\natexlab{a}.
\newblock \showarticletitle{Probabilistically Correct Language-Based
  Multi-Robot Planning Using Conformal Prediction}.
\newblock \bibinfo{journal}{\emph{IEEE Robotics and Automation Letters}}
  (\bibinfo{year}{2024}).
\newblock


\bibitem[\protect\citeauthoryear{Wete, Greenyer, Kudenko, and Nejdl}{Wete
  et~al\mbox{.}}{2024}]%
        {wete2024multi}
\bibfield{author}{\bibinfo{person}{Eric Wete}, \bibinfo{person}{Joel Greenyer},
  \bibinfo{person}{Daniel Kudenko}, {and} \bibinfo{person}{Wolfgang Nejdl}.}
  \bibinfo{year}{2024}\natexlab{}.
\newblock \showarticletitle{Multi-Robot Motion and Task Planning in Automotive
  Production Using Controller-based Safe Reinforcement Learning}. In
  \bibinfo{booktitle}{\emph{Proc. of AAMAS}}.
\newblock


\bibitem[\protect\citeauthoryear{Willick, Eager, and Bunt}{Willick
  et~al\mbox{.}}{1993}]%
        {willick1993disk}
\bibfield{author}{\bibinfo{person}{Darryl~L Willick}, \bibinfo{person}{Derek~L
  Eager}, {and} \bibinfo{person}{Richard~B Bunt}.}
  \bibinfo{year}{1993}\natexlab{}.
\newblock \showarticletitle{Disk cache replacement policies for network
  fileservers}. In \bibinfo{booktitle}{\emph{Proc. of IEEE ICDCS}}.
\newblock


\bibitem[\protect\citeauthoryear{Wu, Guo, Zhang, Sun, Zhang, and Yu}{Wu
  et~al\mbox{.}}{2023}]%
        {wu2023learning}
\bibfield{author}{\bibinfo{person}{Lei Wu}, \bibinfo{person}{Bin Guo},
  \bibinfo{person}{Qiuyun Zhang}, \bibinfo{person}{Zhuo Sun},
  \bibinfo{person}{Jieyi Zhang}, {and} \bibinfo{person}{Zhiwen Yu}.}
  \bibinfo{year}{2023}\natexlab{}.
\newblock \showarticletitle{Learning to Self-Reconfigure for Freeform Modular
  Robots via Altruism Proximal Policy Optimization.}. In
  \bibinfo{booktitle}{\emph{Proc. of IJCAI}}.
\newblock


\bibitem[\protect\citeauthoryear{Xie, Chowdhury, De~Paolis~Kaluza, Zhao, Wong,
  and Yu}{Xie et~al\mbox{.}}{2020}]%
        {xie2020deep}
\bibfield{author}{\bibinfo{person}{Fan Xie}, \bibinfo{person}{Alexander
  Chowdhury}, \bibinfo{person}{M De~Paolis~Kaluza}, \bibinfo{person}{Linfeng
  Zhao}, \bibinfo{person}{Lawson Wong}, {and} \bibinfo{person}{Rose Yu}.}
  \bibinfo{year}{2020}\natexlab{}.
\newblock \showarticletitle{Deep imitation learning for bimanual robotic
  manipulation}. In \bibinfo{booktitle}{\emph{Proc. of NeurIPS}}.
\newblock


\bibitem[\protect\citeauthoryear{Xu, Yu, Wang, Guo, and Han}{Xu
  et~al\mbox{.}}{2019}]%
        {xu2019acousticid}
\bibfield{author}{\bibinfo{person}{Wei Xu}, \bibinfo{person}{ZhiWen Yu},
  \bibinfo{person}{Zhu Wang}, \bibinfo{person}{Bin Guo}, {and}
  \bibinfo{person}{Qi Han}.} \bibinfo{year}{2019}\natexlab{}.
\newblock \showarticletitle{Acousticid: gait-based human identification using
  acoustic signal}.
\newblock \bibinfo{journal}{\emph{Proc. of ACM IMWUT}} (\bibinfo{year}{2019}).
\newblock


\bibitem[\protect\citeauthoryear{Yao, Zhao, Yu, Du, Shafran, Narasimhan, and
  Cao}{Yao et~al\mbox{.}}{2023}]%
        {yao2023react}
\bibfield{author}{\bibinfo{person}{Shunyu Yao}, \bibinfo{person}{Jeffrey Zhao},
  \bibinfo{person}{Dian Yu}, \bibinfo{person}{Nan Du}, \bibinfo{person}{Izhak
  Shafran}, \bibinfo{person}{Karthik Narasimhan}, {and} \bibinfo{person}{Yuan
  Cao}.} \bibinfo{year}{2023}\natexlab{}.
\newblock \showarticletitle{React: Synergizing reasoning and acting in language
  models}. In \bibinfo{booktitle}{\emph{Proc. of ICLR}}.
\newblock


\bibitem[\protect\citeauthoryear{Yu, Yuben, Chao, Zhen, Lei, and Qihui}{Yu
  et~al\mbox{.}}{2025}]%
        {yu2025adaptive}
\bibfield{author}{\bibinfo{person}{LI Yu}, \bibinfo{person}{QU Yuben},
  \bibinfo{person}{DONG Chao}, \bibinfo{person}{QIN Zhen},
  \bibinfo{person}{Zhang Lei}, {and} \bibinfo{person}{WU Qihui}.}
  \bibinfo{year}{2025}\natexlab{}.
\newblock \showarticletitle{Adaptive model switching of collaborative inference
  for multi-CNN streams in UAV swarm}.
\newblock \bibinfo{journal}{\emph{Chinese Journal of Aeronautics}}
  (\bibinfo{year}{2025}).
\newblock


\bibitem[\protect\citeauthoryear{Zhang, Du, Shan, Zhou, Du, Tenenbaum, Shu, and
  Gan}{Zhang et~al\mbox{.}}{2024}]%
        {zhang2023building}
\bibfield{author}{\bibinfo{person}{Hongxin Zhang}, \bibinfo{person}{Weihua Du},
  \bibinfo{person}{Jiaming Shan}, \bibinfo{person}{Qinhong Zhou},
  \bibinfo{person}{Yilun Du}, \bibinfo{person}{Joshua~B Tenenbaum},
  \bibinfo{person}{Tianmin Shu}, {and} \bibinfo{person}{Chuang Gan}.}
  \bibinfo{year}{2024}\natexlab{}.
\newblock \showarticletitle{Building cooperative embodied agents modularly with
  large language models}. In \bibinfo{booktitle}{\emph{Proc. of ICLR}}.
\newblock


\bibitem[\protect\citeauthoryear{Zhang, Yang, and Ba{\c{s}}ar}{Zhang
  et~al\mbox{.}}{2021}]%
        {zhang2021multi}
\bibfield{author}{\bibinfo{person}{Kaiqing Zhang}, \bibinfo{person}{Zhuoran
  Yang}, {and} \bibinfo{person}{Tamer Ba{\c{s}}ar}.}
  \bibinfo{year}{2021}\natexlab{}.
\newblock \showarticletitle{Multi-agent reinforcement learning: A selective
  overview of theories and algorithms}.
\newblock \bibinfo{journal}{\emph{Handbook of reinforcement learning and
  control}} (\bibinfo{year}{2021}).
\newblock


\bibitem[\protect\citeauthoryear{Zhang, Yang, Bai, Wu, Li, Wang, and Li}{Zhang
  et~al\mbox{.}}{2025}]%
        {zhang2024towards}
\bibfield{author}{\bibinfo{person}{Yang Zhang}, \bibinfo{person}{Shixin Yang},
  \bibinfo{person}{Chenjia Bai}, \bibinfo{person}{Fei Wu}, \bibinfo{person}{Xiu
  Li}, \bibinfo{person}{Zhen Wang}, {and} \bibinfo{person}{Xuelong Li}.}
  \bibinfo{year}{2025}\natexlab{}.
\newblock \showarticletitle{Towards efficient llm grounding for embodied
  multi-agent collaboration}.
\newblock \bibinfo{journal}{\emph{Findings of ACL}} (\bibinfo{year}{2025}).
\newblock


\bibitem[\protect\citeauthoryear{Zhao, Lee, and Hsu}{Zhao
  et~al\mbox{.}}{2023}]%
        {zhao2023large}
\bibfield{author}{\bibinfo{person}{Zirui Zhao}, \bibinfo{person}{Wee~Sun Lee},
  {and} \bibinfo{person}{David Hsu}.} \bibinfo{year}{2023}\natexlab{}.
\newblock \showarticletitle{Large language models as commonsense knowledge for
  large-scale task planning}. In \bibinfo{booktitle}{\emph{Proc. of NeurIPS}}.
\newblock


\end{thebibliography}

\appendix
\clearpage
\section*{Appendix}

\vspace{6pt}

\section{MeCoBench} \label{bench}

\begin{figure}[b] 
    \centering
    \begin{subfigure}{0.45\textwidth}
        \centering
        \includegraphics[width=\linewidth]{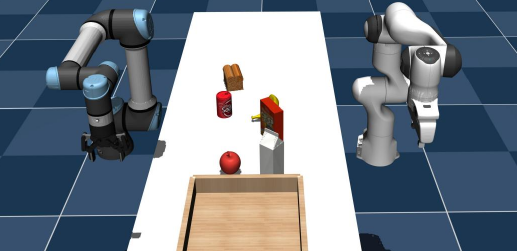}
        \caption{Previous task}
    \end{subfigure}
    \hfill
    \begin{subfigure}{0.45\textwidth}
        \centering
        \includegraphics[width=\linewidth]{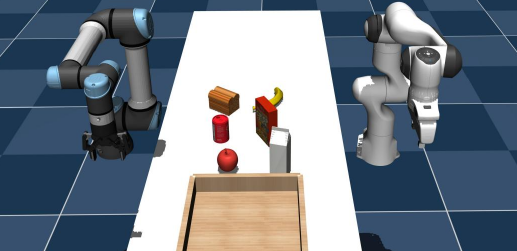}
        \caption{Similar task}
    \end{subfigure}
    \hfill
    \begin{subfigure}{0.45\textwidth}
        \centering
        \includegraphics[width=\linewidth]{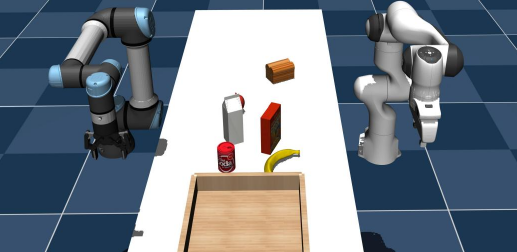}
        \caption{Dissimilar task}
    \end{subfigure}
    \caption{(a) shows previous tasks stored in the task cache, (b) represents similar tasks with an $\alpha$ value greater than 0.8, and (c) illustrates dissimilar tasks with an $\alpha$ value less than 0.8.}
    \label{s_pack}
\end{figure}

\begin{figure}[b] 
    \centering
    \begin{subfigure}{0.45\textwidth}
        \centering
        \includegraphics[width=\linewidth]{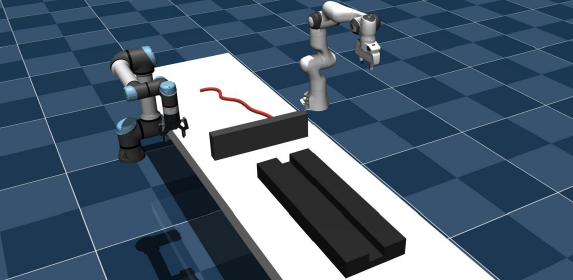}
        \caption{Previous task}
    \end{subfigure}
    \hfill
    \begin{subfigure}{0.45\textwidth}
        \centering
        \includegraphics[width=\linewidth]{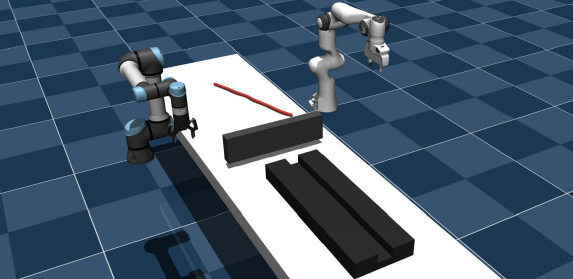}
        \caption{Similar task}
    \end{subfigure}
    \hfill
    \begin{subfigure}{0.45\textwidth}
        \centering
        \includegraphics[width=\linewidth]{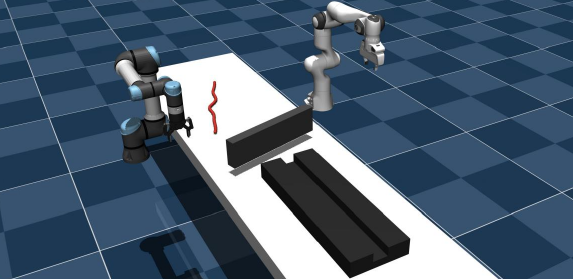}
        \caption{Dissimilar task}
    \end{subfigure}
    \caption{(a) shows previous tasks stored in the task cache, (b) represents similar tasks with an $\alpha$ value greater than 0.7, and (c) illustrates dissimilar tasks with an $\alpha$ value less than 0.7.}
    \label{s_rope}
\end{figure}

\begin{figure}[t] 
    \centering
    \begin{subfigure}{0.465\textwidth}
        \centering
        \includegraphics[width=\linewidth]{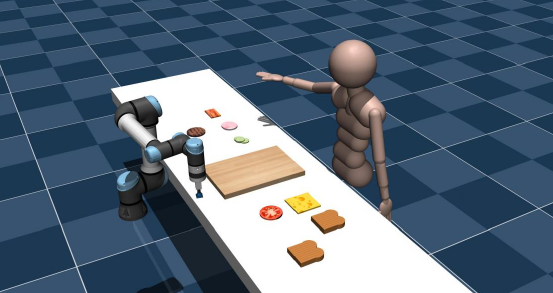}
        \caption{Previous task}
    \end{subfigure}
    \hfill
    \begin{subfigure}{0.465\textwidth}
        \centering
        \includegraphics[width=\linewidth]{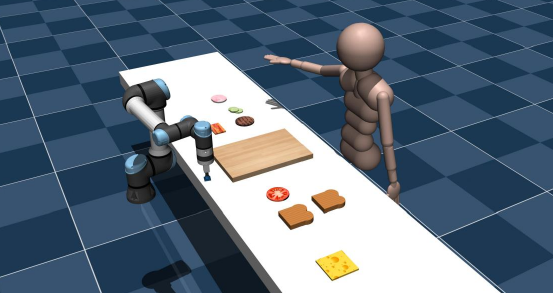}
        \caption{Similar task}
    \end{subfigure}
    \hfill
    \begin{subfigure}{0.465\textwidth}
        \centering
        \includegraphics[width=\linewidth]{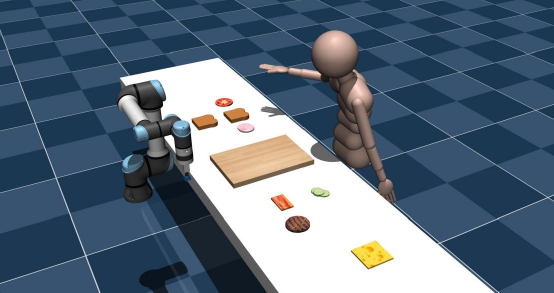}
        \caption{Dissimilar task}
    \end{subfigure}
    \caption{(a) shows previous tasks stored in the task cache, (b) represents similar tasks with the same recipe that satisfy point mapping $\Pi$, and (c) illustrates dissimilar tasks that do not satisfy point mapping $\Pi$.}
    \label{s_sandwich}
\end{figure}

\begin{figure}[t] 
    \centering
    \begin{subfigure}{0.42\textwidth}
        \centering
        \includegraphics[width=\linewidth]{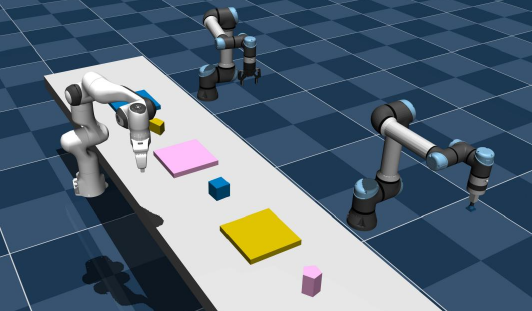}
        \caption{Previous task}
    \end{subfigure}
    \hfill
    \begin{subfigure}{0.42\textwidth}
        \centering
        \includegraphics[width=\linewidth]{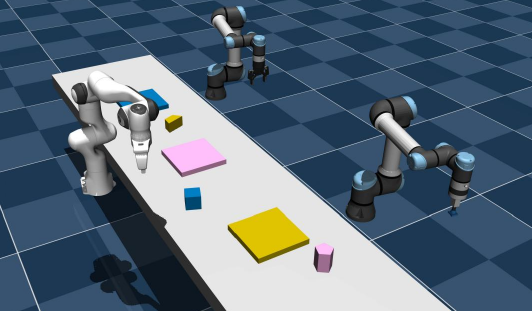}
        \caption{Similar task}
    \end{subfigure}
    \hfill
    \begin{subfigure}{0.42\textwidth}
        \centering
        \includegraphics[width=\linewidth]{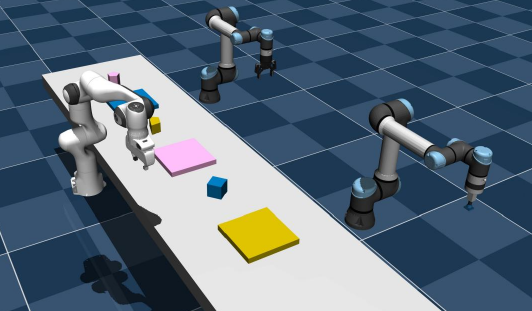}
        \caption{Dissimilar task}
    \end{subfigure}
    \caption{(a) shows previous tasks stored in the task cache, (b) represents similar tasks that satisfy point mapping $\Pi$, and (c) illustrates dissimilar tasks that do not satisfy point mapping~$\Pi$.}
    \label{s_sort}
\end{figure}

We present a variant of RoCoBench~\cite{mandi2024roco} for similar tasks, called MeCoBench. MeCoBench includes six tasks: \textit{Pack Grocery}, \textit{Move Rope}, \textit{Make Sandwich}, \textit{Sort Cubes}, \textit{Sweep Floor}, and \textit{Arrange Cabinet}. These tasks are categorized into two groups: high-workspace-overlap tasks and low-workspace-overlap tasks. \textit{Pack Grocery} and \textit{Move Rope} belong to the high-workspace-overlap category, while \textit{Make Sandwich}, \textit{Sort Cubes}, \textit{Sweep Floor}, and \textit{Arrange Cabinet} fall under the low-workspace-overlap category. We use $\alpha$ and $\Pi$ to measure task similarity for the high and low workspace overlap groups, respectively, as defined in Section~\ref{sec:A}. This allows MeCoBench to operate in three modes: randomly generating tasks that are \mbox{\textbf{similar}} to those in the task cache, \textbf{dissimilar} to them, and \mbox{\textbf{without considering similarity at all}}.

\subsection{Pack Grocery} 
\textbf{Task Description.} Two robots are positioned on opposite sides of a table: the UR5E robot (`Alice') and the Franka Panda (`Bob'). On the table, there are six distinct items (apple, banana, milk, soda can, bread, and cereal) and a fixed-position box. The robots are tasked with placing the items into the box, with the items' positions randomly generated.

\textbf{Similarity setting.} As shown in Figure~\ref{s_pack}, we set the $\alpha$ threshold to 0.85. Tasks with an $\alpha$ value below 0.85 are considered dissimilar; otherwise, they are considered similar.

\subsection{Move Rope} 
\textbf{Task Description.} Two robots are positioned on opposite sides of a table: the UR5E robot (`Alice') and the Franka Panda (`Bob'). A rope, a groove, and a wall are placed between them. The robots must grasp the ends of the rope and place them into the groove. The groove has a fixed position, while the positions of the rope and the wall are randomly generated within a predefined range.

\textbf{Similarity setting.} As shown in Figure~\ref{s_rope}, we set the $\alpha$ threshold to 0.7. Tasks with an $\alpha$ value below 0.7 are considered dissimilar; otherwise, they are considered similar.

\subsection{Make Sandwich} 
\textbf{Task Description.} Two robots are positioned on opposite sides of a table: a UR5E robot (`Chad') and a humanoid robot (`Dave'). A cutting board is placed at a fixed position on the table, with eight food items randomly distributed on either side. The left side of the board can hold up to six items, while the right side can hold up to eight. The robots are tasked with preparing sandwiches based on one of four predefined recipes. Chad is restricted to retrieving food from the right side of the board, and Dave is also limited to retrieving food from the left side.

\textbf{Similarity setting.} The similarity between two tasks is primarily determined by matching their recipes. Additionally, as shown in Figure~\ref{s_sandwich}, if the tasks satisfy point mapping $\Pi$, they are considered similar. In this task, the cutting board divides the table into two regions: the left and right sides. Therefore, the task space $\Omega$ is partitioned into three regions: $A_1$ on the left side of the cutting board, $A_2$ on the cutting board, and $A_3$ on the left side of the cutting~board.

\subsection{Sort Cubes} 
\textbf{Task Description.} Three robots are positioned on either side of the table: the UR5E robot (`Alice'), the Franka Panda (`Bob'), and another UR5E robot (`Chad'). On the table, there are three blocks in different colors, along with three fixed panels that correspond to each block's color. These panels divide the table into four regions. Each block is randomly placed in one of the regions, with its exact position within the region also determined randomly. The robots must place the blocks onto the corresponding panels using their limited range of motion.

\textbf{Similarity setting.} As shown in Figure~\ref{s_sort}, If a task satisfies point mapping $\Pi$, it is considered similar. In this task, the three colored panels divide the table into four regions. Therefore, the task space $\Omega$ is partitioned into seven regions in this case.

\subsection{Sweep Floor} 
\textbf{Task Description.} Two robots are positioned on opposite sides of a table: the UR5E robot (`Alice') and the Franka Panda (`Bob'). Each robot is equipped with a dustpan and a broom. The table contains three blocks and a fixed-position trash bin, with the blocks' positions randomly generated. The two robots must collaborate to sweep the blocks into the trash bin.

\begin{figure}[t] 
    \centering
    \begin{subfigure}{0.45\textwidth}
        \centering
        \includegraphics[width=\linewidth]{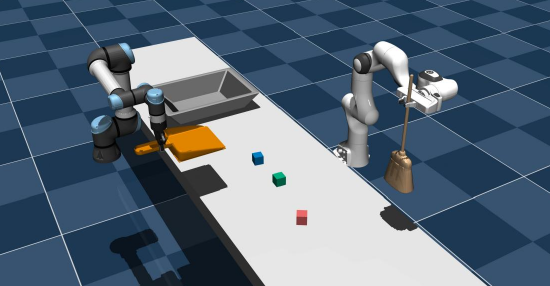}
        \caption{Previous task}
    \end{subfigure}
    \hfill
    \begin{subfigure}{0.45\textwidth}
        \centering
        \includegraphics[width=\linewidth]{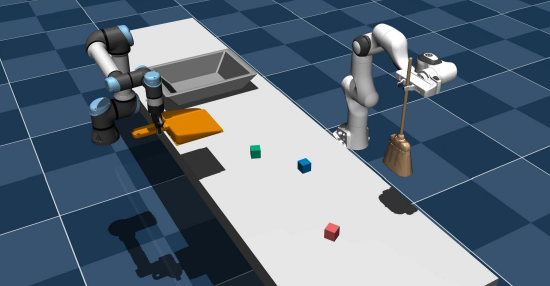}
        \caption{Similar task}
    \end{subfigure}
    \hfill
    \caption{(a) shows previous tasks stored in the task cache, (b) represents similar tasks that satisfy point mapping $\Pi$.}
    \label{s_sweep}
\end{figure}

\textbf{Similarity setting.} As shown in Figure~\ref{s_sweep}, in this task, the task space $\Omega$ cannot be further partitioned. In this case, $\mathcal{A} = {A_1}$, where $A_1$ represents the table. Therefore, the tasks are highly similar to each other, and there are no cases where two tasks are dissimilar.

\subsection{Arrange Cabinet} 
\textbf{Task Description.} Three robots are positioned on either side of the table: a UR5E robot (`Alice'), a Franka Panda (`Bob'), and another UR5E robot (`Chad'). A two-level cabinet is placed on the table, containing a mug and a cup. The cabinet remains in a fixed position, while the mug and cup are randomly placed inside it. The robots must collaborate to open the cabinet door and place the mug and cup onto their corresponding coasters.

\begin{figure}[t] 
    \centering
    \begin{subfigure}{0.395\textwidth}
        \centering
        \includegraphics[width=\linewidth]{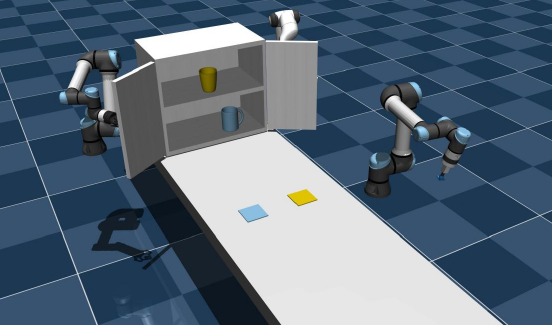}
        \caption{Previous task}
    \end{subfigure}
    \hfill
    \begin{subfigure}{0.395\textwidth}
        \centering
        \includegraphics[width=\linewidth]{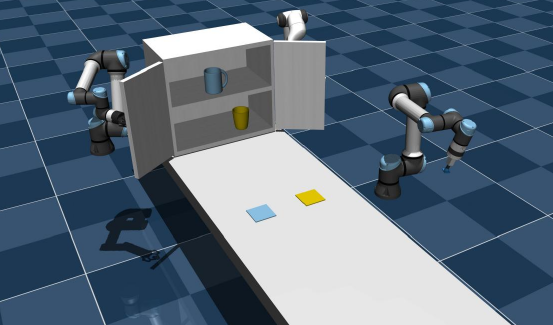}
        \caption{Similar task}
    \end{subfigure}
    \hfill
    \caption{(a) shows previous tasks stored in the task cache, (b) represents similar tasks that satisfy point mapping $\Pi$.}
    \label{s_cabinet}
\end{figure}

\textbf{Similarity setting.} As shown in Figure~\ref{s_cabinet}, in this task, the task space $\Omega$ is partitioned into two regions: $A_1$ inside the cabinet and $A_2$ on the table outside the cabinet. Therefore, the tasks are highly similar to each other, and there are no cases where~two tasks~are~dissimilar.

\begin{figure*}[!t]
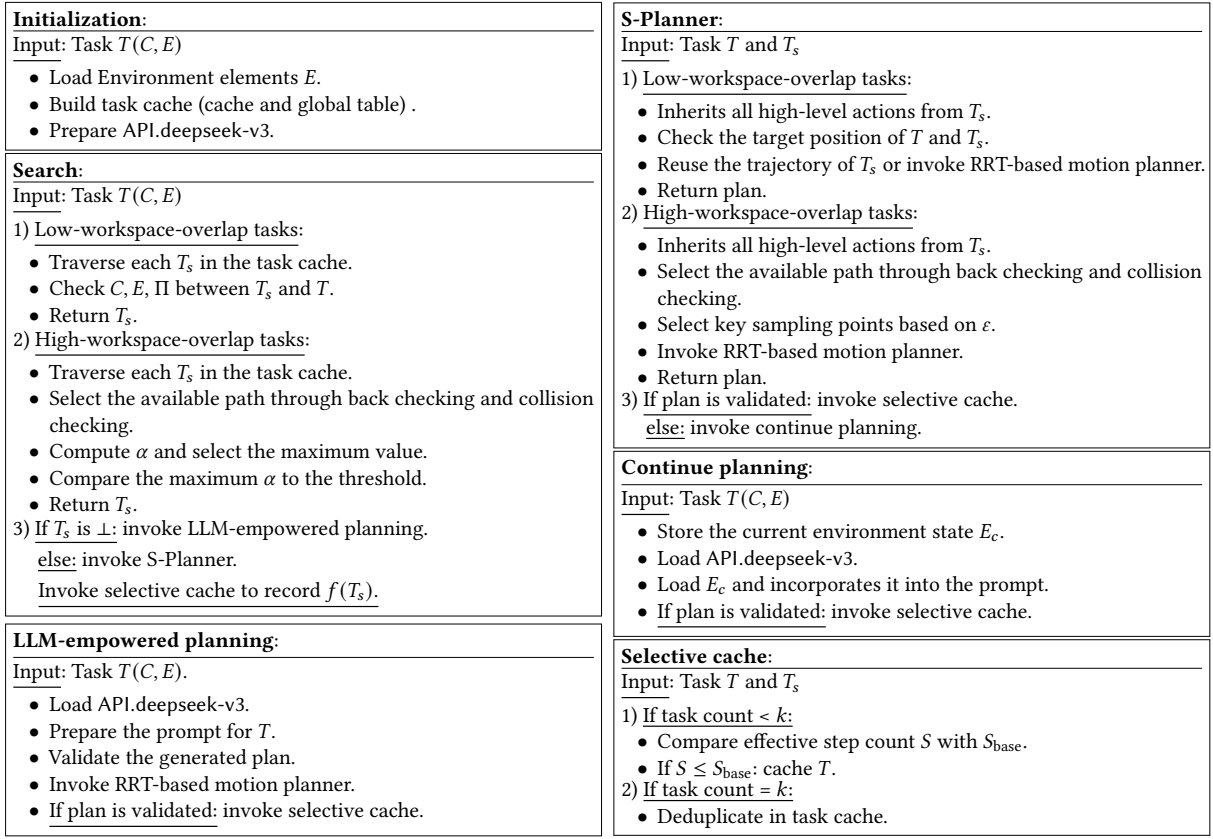

\centering\resizebox{0.45\textwidth}{!}{
\begin{minipage}[!t]{.49\linewidth}
\begin{boxedminipage}[H]{\linewidth}
\textbf{Initialization}:
\smallskip
\hrule
\smallskip

\underline{Input}: Task \( T (C, E) \) 

\smallskip
\begin{itemize}[leftmargin=15pt, itemsep=0pt, topsep=-3pt,parsep=0pt]
\item Load Environment elements \( E \).
\item Build task cache (cache and global table) .
\item Prepare $\mathsf{API}.\mathsf{deepseek\text{-}v3}$.
\end{itemize}
\end{boxedminipage}

\begin{boxedminipage}[H]{\linewidth}
\textbf{Search}:
\smallskip
\hrule
\smallskip
\RestyleAlgo{plain}
\SetAlgoShortEnd

\underline{Input}: Task \( T (C, E) \)
\smallskip

1) \underline{Low\text{-}workspace\text{-}overlap tasks}:

\smallskip

\begin{itemize}[leftmargin=15pt, itemsep=0pt, topsep=-3pt,parsep=0pt]
\item Traverse each \( T_s \) in the task cache.
\item Check \( C, E \), $\Pi$ between \( T_s \) and \( T \).
\item Return \( T_s \).
\end{itemize}

\smallskip

2) \underline{High\text{-}workspace\text{-}overlap tasks}:

\smallskip

\begin{itemize}[leftmargin=15pt, itemsep=0pt, topsep=-3pt,parsep=0pt]
\item Traverse each \( T_s \) in the task cache.
\item Select the available path through back checking and collision checking.
\item Compute $\alpha$ and select the maximum value.
\item Compare the maximum $\alpha$ to the threshold.
\item Return \( T_s \).
\end{itemize}

\smallskip

3) \underline{If \( T_s \) is $\bot$:} invoke LLM-empowered planning.

\smallskip

\hspace{2.8mm} \underline{else:} invoke S-Planner.

\smallskip

\hspace{2.9mm} \underline{Invoke selective cache to record \( f(T_s) \).} 

\end{boxedminipage}

\smallskip

\begin{boxedminipage}[H]{\linewidth}
\textbf{LLM-empowered planning}:
\smallskip
\hrule
\smallskip

\underline{Input}: Task \( T (C, E) \).

\smallskip

\begin{itemize}[leftmargin=15pt, itemsep=0pt, topsep=-3pt,parsep=0pt]
\item Load $\mathsf{API}.\mathsf{deepseek\text{-}v3}$.
\item Prepare the prompt for \( T \).
\item Validate the generated plan.
\item Invoke RRT-based motion planner.
\item \underline{If plan is validated:} invoke selective cache.
\end{itemize}

\end{boxedminipage}
\end{minipage}}
\resizebox{0.45\textwidth}{!}{
\begin{minipage}[!t]{.49\linewidth}

\begin{boxedminipage}[H]{\linewidth}
\textbf{S-Planner}:
\smallskip
\hrule
\smallskip
\RestyleAlgo{plain}
\SetAlgoShortEnd

\underline{Input}: Task \( T  \) and \( T_s  \)
\smallskip

1) \underline{Low\text{-}workspace\text{-}overlap tasks}:

\smallskip

\begin{itemize}[leftmargin=15pt, itemsep=0pt, topsep=-4pt,parsep=0pt]
\item Inherits all high\text{-}level actions from \( T_s  \).
\item Check the target position of \( T  \) and \( T_s  \).
\item Reuse the trajectory of \( T_s  \) or invoke RRT-based motion planner.
\item Return plan.
\end{itemize}

\smallskip

2) \underline{High\text{-}workspace\text{-}overlap tasks}:

\smallskip

\begin{itemize}[leftmargin=15pt, itemsep=0pt, topsep=-4pt,parsep=0pt]
\item Inherits all high\text{-}level actions from \( T_s  \).
\item Select the available path through back checking and collision checking.
\item Select  key sampling points based on \( \varepsilon \).
\item Invoke RRT-based motion planner.
\item Return plan.
\end{itemize}

\smallskip

3) \underline{If plan is validated:} invoke selective cache.

\hspace{2.8mm} \underline{else:} invoke continue planning.

\end{boxedminipage}

\begin{boxedminipage}[H]{\linewidth}
\textbf{Continue planning}:
\smallskip
\hrule
\smallskip

\underline{Input}: Task \( T (C, E) \) 

\smallskip
\begin{itemize}[leftmargin=15pt, itemsep=0pt, topsep=-4pt,parsep=0pt]
\item Store the current environment state \( E_c \).
\item Load $\mathsf{API}.\mathsf{deepseek\text{-}v3}$.
\item Load \( E_c \) and incorporates it into the prompt.
\item \underline{If plan is validated:} invoke selective cache.
\end{itemize}
\end{boxedminipage}

\begin{boxedminipage}[H]{\linewidth}
\textbf{Selective cache}:
\smallskip
\hrule
\smallskip
\RestyleAlgo{plain}
\SetAlgoShortEnd

\underline{Input}: Task \( T  \) and \( T_s  \)
\smallskip

1) \underline{If task count < $k$:}

\begin{itemize}[leftmargin=15pt, itemsep=0pt, topsep=-4pt,parsep=0pt]
\item Compare effective step count \( S \) with  \(S_{\text{base}}\).
\item If \( S \leq S_{\text{base}} \): cache \( T  \).
\end{itemize}

\smallskip

2) \underline{If task count = $k$:}

\begin{itemize}[leftmargin=15pt, itemsep=0pt, topsep=-4pt,parsep=0pt]
\item Deduplicate in task cache. 
\end{itemize}

\end{boxedminipage}
\end{minipage}}
\caption{The basic workflow of MeCo.}
\label{wrapping up}
\end{figure*}

\section{Detailed workflow of MeCo}

\begin{figure*}[h] 
    \centering
    \begin{subfigure}{0.9\textwidth}
        \centering
        \includegraphics[width=\linewidth]{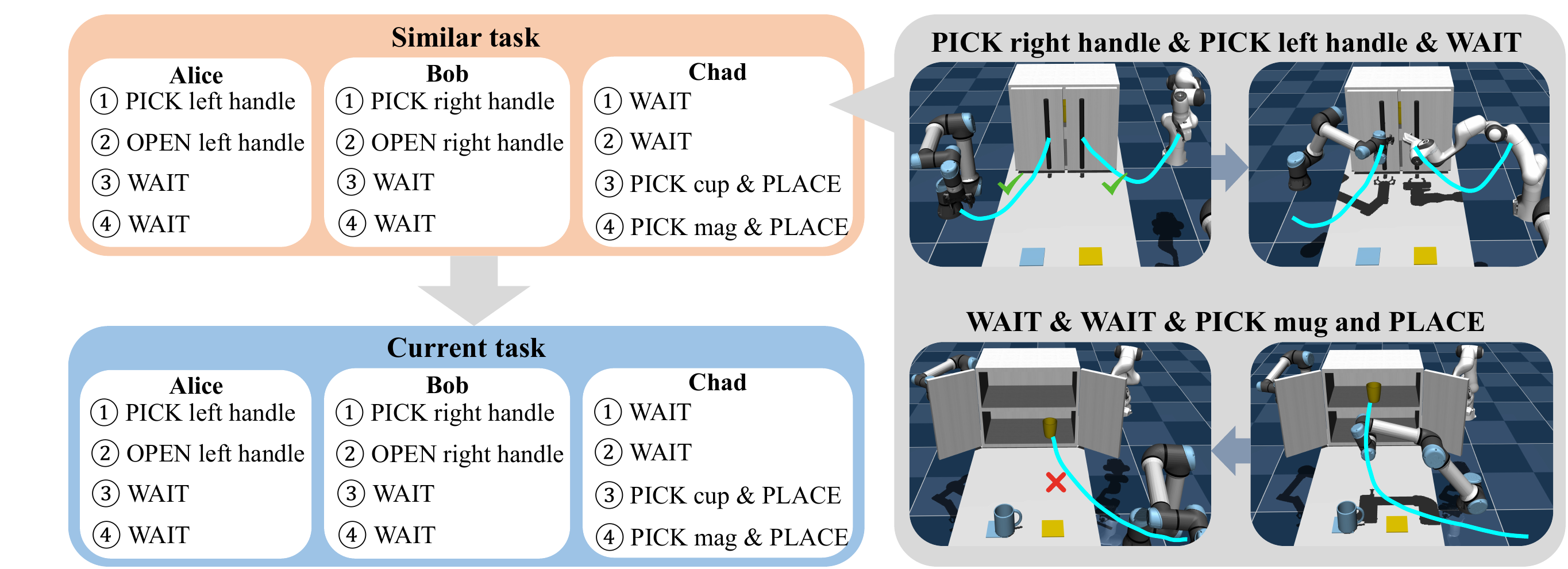}
    \end{subfigure}
    \caption{Workflow of S-Planner in the low-workspace-overlap mode. 
    The left side illustrates how tasks are decomposed and assigned at the high level, 
    while the right side shows how motion planning is conducted at the low level. 
    The blue curves represent the reference trajectory.
    `Alice', `Bob' and `Chad' refer to robots.}
    \label{S-Planner_1}
\end{figure*}

As shown in Figure~\ref{wrapping up}, we now conclude by presenting the overall workflow of MeCo. 
After each successful planning, MeCo selectively stores the task plan in the task cache for future reference. When planning a new task, it first searches the task cache for a similar task that meets the current requirements. If no similar task is found, MeCo calls LLMs and uses the task description as the prompt for planning. If a similar task is matched, it further invokes the similar motion planner (S-Planner), which generates a plan for the current task efficiently by referencing the similar task without relying on LLMs. 
S-Planner or LLMs decompose the task into multiple subtasks. For each subtask, they output 3D waypoint paths to guide the low-level motion planner. 
The generated plan undergoes a set of validation, such as inverse kinematics (IK) and collision checking. If validation fails, the failure reason is fed back to LLMs for replanning. Specifically, if S-Planner fails to generate a valid plan, LLMs continue planning from the failed step instead of restarting from scratch. 
Once all validations are passed, the final motion trajectories for all robots are planned by the low-level motion planner and executed in the environment.
The detailed workflow of each step is described as follows.

\begin{figure}[t]
    \centering
    \begin{subfigure}{0.48\textwidth}
        \centering
        \includegraphics[width=\linewidth]{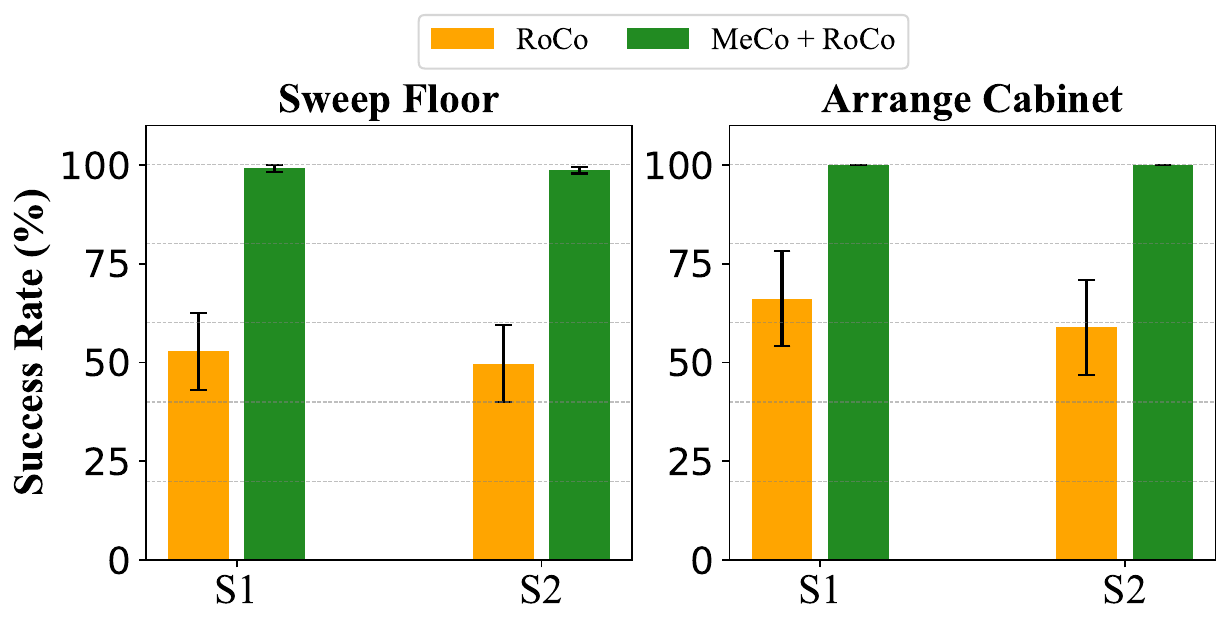}
        \label{ablation_t}
    \end{subfigure}

    \vspace{-9pt}

    \begin{subfigure}{0.48\textwidth}
        \centering
        \includegraphics[width=\linewidth]{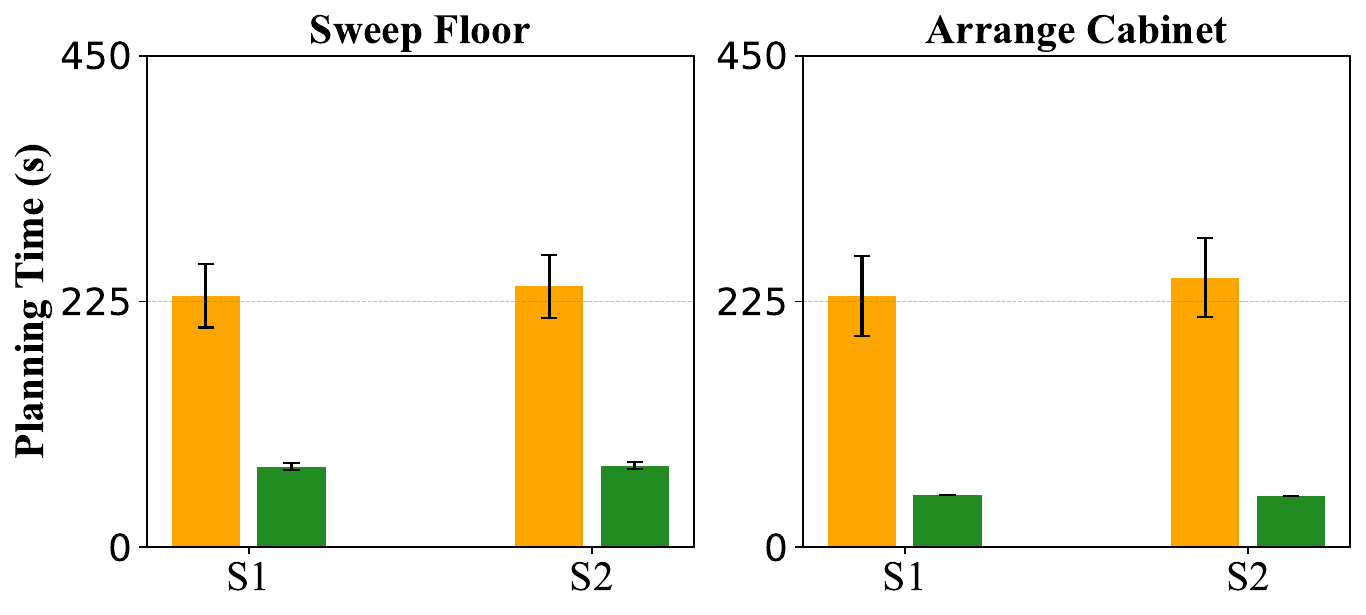}
        \label{ablation_p}
    \end{subfigure}

    \vspace{-9pt}

    \begin{subfigure}{0.48\textwidth}
        \centering
        \includegraphics[width=\linewidth]{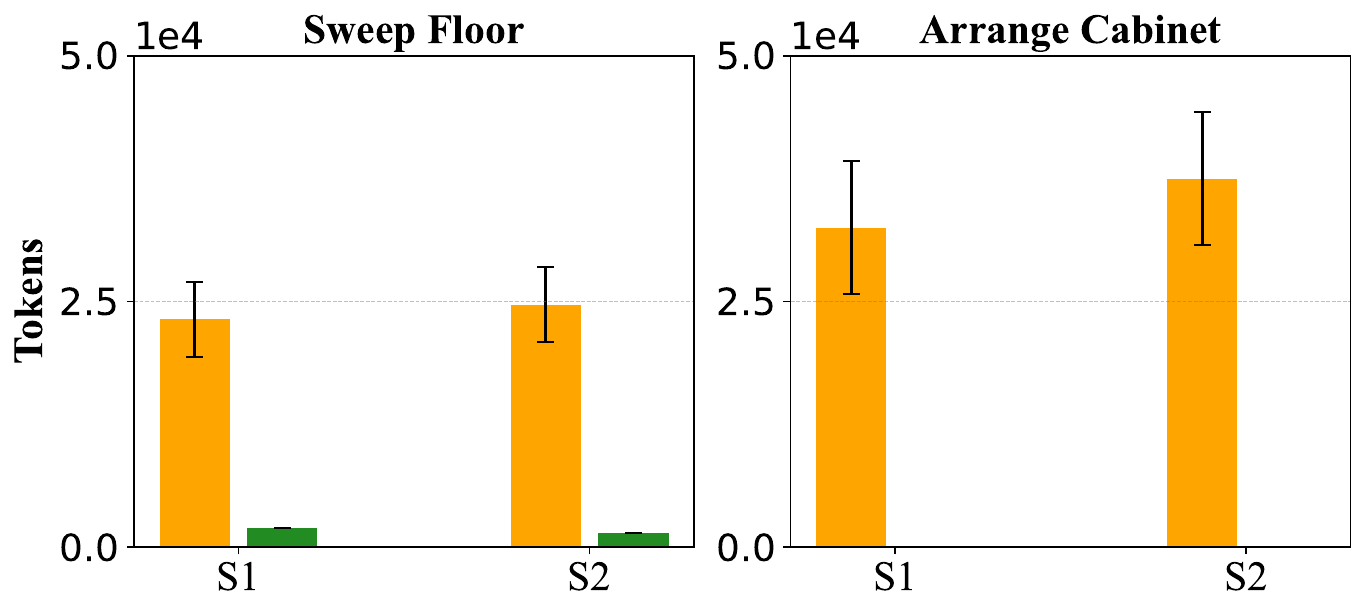}
        \label{ablation_t}
    \end{subfigure}

    \caption{The performance of MeCo on \textit{Sweep Floor} and \textit{Arrange Cabinet}. We evaluate on MeCoBench across three scenarios: random (S1), totally similar (S2), and totally different (S3), using RoCo~\cite{mandi2024roco} for comparison. However, for the \textit{Sweep Floor} and \textit{Arrange Cabinet} tasks, there are no totally different (S3) scenarios, as the tasks are highly similar. For each task, we report success rate, planning time, and token consumption averaged over 30 random seeds.}
    \label{ablation_a}

\end{figure}

\textbf{Initialization.} When the system receives a new task \( T \), it first loads the task environment \( E \). Next, it creates a task cache, which includes a cache table for speeding up processes and a global table for frequency updates. After that, the deepseek-v3 API is prepared for the subsequent steps.

\textbf{Search.} For low-workspace-overlap tasks, we iterate through each task $T_s$ in the task cache and check its compatibility with the current task $T$ in terms of $C$, $E$, and whether they satisfy the point mapping $\Pi$.
For high-workspace-overlap tasks, iterate through each task \( T_s \) and select the available path based on the \( E \). Then calculate the $\alpha$ for each \( T_s \) and compare the max $\alpha$ to the threshold. If no matching tasks, invoke LLM-empowered planning; Otherwise, invoke S-Planner. Finally, call the selective cache to record \( f(T_s) \).

\textbf{LLM-empowered planning.} The system loads the deepseek-v3 API and then prepares the prompt based on the \( E \) of the current task to provide input for LLMs planning. The generated plan undergoes a series of validations, and if it passes, it is executed and calls the selective cache.

\textbf{S-Planner.} For low-workspace-overlap tasks, as shown in Figure~\ref{S-Planner_1}, the current task \( T \) inherits the high-level actions from~\( T_s \). Then checks if the target positions are the same to either reuse or replan the trajectory. For high-workspace-overlap tasks, \( T \) inherits the high-level actions from \( T_s \). Then, it determines the \( R_{\text{final}} \) through back and collision checking, and selects key sampling points for trajectory planning. If the plan is validated, selective cache is invoked; Otherwise, continuous planning is used.

\begin{figure}[t]
    \makebox[\columnwidth][c]{
    \includegraphics[width=1\columnwidth]{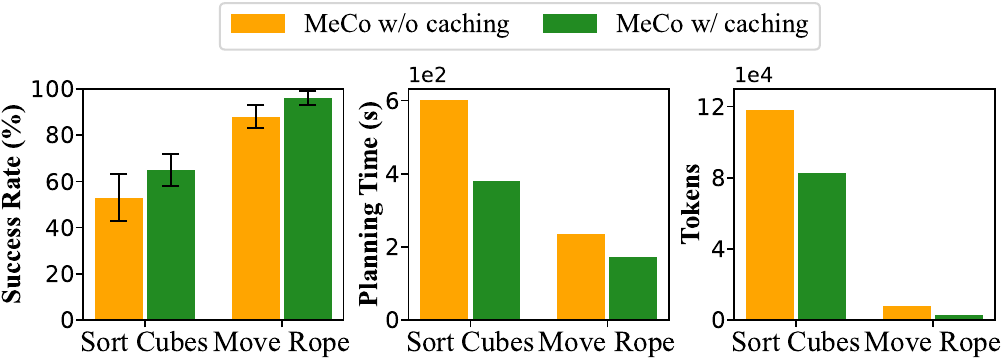}
    }
    \caption{Ablation study on the selective caching mechanism. To accurately evaluate the importance of the selective caching mechanism, we store identical tasks in the cache and test each task using more than 50 random seeds on average.}
    \label{ablation}
\end{figure}

\section{Additional Experiments}

\subsection{Performance of MeCo} \label{A1}

As shown in Figure~\ref{ablation_a}, we evaluate the effectiveness of our method on the \textit{Sweep Floor} and \textit{Arrange Cabinet} tasks. In terms of success rate, \textit{Sweep Floor} achieves a 46\% improvement and \textit{Arrange Cabinet} achieves a 34\% improvement, with both tasks reaching or approaching 100\% success. Regarding planning time, \textit{Sweep Floor} is reduced by 68\% and \textit{Arrange Cabinet} by 79\%. For token consumption, \textit{Sweep Floor} decreases by 91\% and \textit{Arrange Cabinet} by 100\%. These results demonstrate the effectiveness of our method, which significantly reduces planning costs while improving task success rates.

\subsection{Ablation studies} \label{A2}

To verify the importance of the selective caching mechanism, we remove the selective caching mechanism, allowing all successfully tasks to be stored without filtering or deduplication until the cache limit is reached. 
Specifically, we simulate a case where the cache contains identical tasks. In this situation, the cache exhibits extremely high redundancy, resulting in very low task diversity. We then select two representative tasks, \textit{Move Rope} and \textit{Sort Cubes}, for~evaluation. 

As shown in Figure~\ref{ablation}, 
the impact of the selective caching mechanism is evident: the success rate increases by an average of 33\%, the planning time decreases by 32\%, and the token consumption drops by 47\%. These results demonstrate that the selective caching mechanism enhances the diversity of cached tasks, enabling MeCo to handle more diverse scenarios.

\section{Discussion}

\subsection{System complexity and thresholds} 

MeCo introduces several components on top of an LLM-empowered planner, including similarity testing, S-Planner, continuous planning, and selective caching. These modules inevitably increase system complexity. However, when tasks are highly similar, they deliver substantial benefits. Experiments show that MeCo improves success rate while reducing planning time and token usage, which supports the value and feasibility of these design choices.

Regarding the hyperparameters, MeCo introduces four parameters. They are conveniently determined offline for each task category and remain stable during system operation.

\begin{itemize}[leftmargin=*]
    \item \textbf{Tolerance threshold $\varepsilon$:}
    For low-workspace-overlap tasks, the S-Planner reuses the historical trajectory only when the target-pose difference stays within $\varepsilon$. The $\varepsilon$ typically does not require frequent tuning.

    \item \textbf{Expansion coefficient $\eta$:}
    For high-workspace-overlap tasks, the S-Planner selects key sampling points from a validated reference trajectory based on $\eta$. It usually remains stable because it mainly depends on trajectory discretization and sparsity of key points.

    \item \textbf{Similarity threshold $\tau$:}
    This threshold controls how strict similar-task matching is. A lower value relaxes the match and encourages trying more historical tasks, while a higher value increases the probability that the chosen reference trajectory is reusable.

    \item \textbf{Cache size $k$:}
    Cache size controls search overhead and the diversity of stored tasks. A smaller $k$ reduces latency, whereas a larger $k$ often improves success rate and further reduces token usage.
\end{itemize}

\subsection{Zero-shot environments} 
In zero-shot environments, MeCo leverages the ability of LLMs to capture common-sense knowledge for flexible task decision-making. Moreover, MeCo is essentially orthogonal to the underlying LLM implementation and does not rely on the internal implementation or reasoning mechanisms. Therefore, MeCo can easily integrate with existing LLM-based multi-agent frameworks such as HMAS-2~\cite{chen2024scalable}, RoCo~\cite{mandi2024roco}, and ReAd~\cite{zhang2024towards}.

However, MeCo is not limited to LLM-based planners. Since the interaction operates as a black box, MeCo can potentially extend to other types of planners, even to manually designed ones, when LLM is unreliable. It could be helpful to address zero-shot environments.

\subsection{Unstructured scenes}

In unstructured environments, the low-workspace-overlap assumption, where the workspace can be cleanly partitioned into multiple regions, is often hard to satisfy. 
This undermines region-based similarity criteria that assume consistent cross-region interaction patterns. In addition, unstructured scenes typically increase inter-robot interference, making collision risks more frequent.

In this case, we can treat the problem as a high-workspace-overlap task and use the reuse ratio $\alpha$ for similarity testing and subsequent planning. This is because $\alpha$ 
 is determined according to reusable trajectory segments and is therefore not affected by the unstructuredness of the scene.

\section{Limitation and future work}
We recognise several limitations in our work. First, the threshold $\alpha$ for identifying similar tasks and the maximum number of cached tasks $k$ are manually tuned through experiments. These parameters need to be optimized separately for different tasks, which limits the system's adaptability to new tasks. Future efforts could explore strategies such as self-supervised or meta-learning to automatically learn optimal parameters. Second, MeCo currently relies mainly on textual descriptions and spatial information. Future work may incorporate multimodal inputs and feedback, such as visual and auditory cues, to improve contextual task understanding. Third, the current task selection and update mechanism is based on frequency. Future work could design a more comprehensive caching strategy by considering factors like task complexity or planning costs, 
thereby improving the representativeness and utility of cached~tasks.

Finally, MeCo focuses on several representative collaborative multi-arm tasks, as multi-arm systems constitute a prototypical form of multi-robot collaboration. However, our method is not restricted to multi-arm settings and can be extended to scenarios involving other types of robots.
Future work can further exploit task similarity to address and optimize more complex tasks. Moreover, task-level similarity can be explored to improve the communication efficiency and reduce the interaction costs between agents and humans in human–robot interaction, such as in frameworks like BOTH~\cite{wang2025both} and hmOS~\cite{wang2024hmos}. In addition, the use of similarity can be extended to various domains of multi-agent collaboration, such as mobile robot navigation and heterogeneous robot cooperation.
For example, for mobile robots, we can first guide the robot to fixed locations using standard navigation methods and then apply our similarity metrics.

\end{document}